\documentclass{article}

\usepackage{microtype}
\usepackage{graphicx}
\usepackage{subfigure}
\usepackage{booktabs} % for professional tables

\usepackage{hyperref}

\usepackage[accepted]{icml2024}

\usepackage{amsmath}
\usepackage{amssymb}
\usepackage{mathtools}
\usepackage{amsthm}

\usepackage[capitalize,noabbrev]{cleveref}

\theoremstyle{plain}

\theoremstyle{definition}

\theoremstyle{remark}

\icmltitlerunning{Surprisal Driven $k$-NN for Robust and Interpretable Nonparametric Learning}

\begin{document}

\twocolumn[
\icmltitle{Surprisal Driven $k$-NN for Robust and Interpretable Nonparametric Learning}



% \icmlsetsymbol{first}{*}

\begin{icmlauthorlist}
\icmlauthor{Amartya Banerjee}{first}
\icmlauthor{Christopher J. Hazard}{howso}
\icmlauthor{Jacob Beel}{howso}
\icmlauthor{Cade Mack}{howso}
\icmlauthor{Jack Xia}{howso}
\icmlauthor{Michael Resnick}{howso}
\icmlauthor{Will Goddin}{howso}
\end{icmlauthorlist}

\icmlaffiliation{first}{Department of Computer Science, UNC-Chapel Hill, NC, USA}
\icmlaffiliation{howso}{Howso Incorporated}

\icmlcorrespondingauthor{Howso Incorporated}{info@howso.com}

\icmlkeywords{Machine Learning, ICML}

\vskip 0.3in
]



\printAffiliationsAndNotice{}  % leave blank if no need to mention equal contribution

\begin{abstract}
Nonparametric learning is a fundamental concept in machine learning that aims to capture complex patterns and relationships in data without making strong assumptions about the underlying data distribution. Owing to simplicity and familiarity, one of the most well-known algorithms under this paradigm is the $k$-nearest neighbors ($k$-NN) algorithm. Driven by the usage of machine learning in safety-critical applications, in this work, we shed new light on the traditional nearest neighbors algorithm from the perspective of information theory and propose a robust and interpretable framework for tasks such as classification, regression, density estimation, and anomaly detection using a single model. We can determine data point weights as well as feature contributions by calculating the conditional entropy for adding a feature without the need for explicit model training. This allows us to compute feature contributions by providing detailed data point influence weights with perfect attribution and can be used to query counterfactuals. Instead of using a traditional distance measure which needs to be scaled and contextualized, we use a novel formulation of \textit{surprisal} (amount of information required to explain the difference between the observed and expected result). 
Finally, our work showcases the architecture's versatility by achieving state-of-the-art results in classification and anomaly detection, while also attaining competitive results for regression across a statistically significant number of datasets.
\end{abstract}

\section{Introduction}
\label{sec:intro}
Nonparametric methods, such as $k$-Nearest Neighbors ($k$-NN), have been studied and applied in various domains 
of statistics and machine learning. Unlike parametric models, nonparametric methods do not rely on a fixed 
number of parameters or make strict distributional assumptions about the underlying data.  This allows for 
algorithms to flexibly adapt to different types of data and capture intricate structures.  First proposed by 
\cite{fix_hodges_fix1951discriminatory}, and \cite{cover_knn}, $k$-NN has seen several modifications and evolutions over the past decades \cite{instanceBased_knnAhaKA91,knn_wang-kulkarni-verdu,esml_book,knn_voting_alpaydin}.  
Despite these advancements, $k$-NN still has some disadvantages. For example, the curse of dimensionality~\cite{esml_book,
IndykApprox,SchuhImproving,TaoQuality}, the selection of a distance metric~\cite{SuryaDist}, and imbalanced datasets~\cite{imbalance5128907} all present significant challenges to $k$-NN.

In this work, we propose methods to enhance $k$-NN to address these issues, derive new concepts driven by entropy, and then demonstrate the performance of this enhanced $k$-NN on various applications. Using our methodology, we are able to improve the performance of $k$-NN while retaining its natural interpretability. Additionally, these improvements allow us to understand the importance of features and weigh them 
accordingly in the model's decision making, thereby improving interpretability further.  First, we derive the \textit{Łukaszyk–Karmowski} 
(LK) distance \cite{lk_phdthesis, lk_article} for Laplace distributions to prevent distances of zero based on uncertainty.  To our knowledge, this is the first 
published derivation of that result. Second, we show how Inverse Residual Weighting (IRW) can be used to move our distance measurements into surprisal space.  Then, we introduce the concept of \textit{conviction}: a ratio of expected surprisal to observed surprisal. This is further broken down into \textit{familiarity conviction}, \textit{similarity conviction}, and \textit{residual conviction}.  Finally, 
we show how these methods and concepts can be used to achieve near or above state-of-the-art results on classification, regression, and anomaly detection tasks.

\section{Related Work}~\label{sec:related}% \subsection{Nearest Neighbors}
In recent years, $k$-NN based methods have grown in popularity in Natural Language Processing (NLP) and Computer Vision (CV). In NLP, variants of $k$-NN have been used on machine translation tasks \cite{knn_iclr_khandelwal2021nearest}, \cite{acl_knn_jiang-etal-2022-towards}, \cite{acl_mt_meng-etal-2022-fast}. 
In \cite{knn_iclr_khandelwal2021nearest}, the authors propose carrying out translation using a large database of pre-translated sentences or phrases as a reference, and during translation, the system searches for the most similar sentences in the database as the translation output. \cite{acl_mt_meng-etal-2022-fast} built on top of this work by proposing an efficient indexing scheme to organize the reference database, enabling faster search and retrieval of the nearest neighbors. This indexing scheme reduces the computational complexity of the translation process and improves overall efficiency. Another line of work that has gained attention in the recent past focuses on low resource text classification using $k$-NN on compressed text data \cite{gzip_knn}.
Other methods use $k$-NN as an auxiliary model on intermediate representations of neural networks for filtering samples.  \cite{bahri2020deepknn} proposed a version of the $k$-NN algorithm called Deep $k$-NN. This algorithm incorporates the principles of $k$-NN into deep learning architectures, specifically convolutional neural networks (CNNs), to effectively handle noisy labels. 
The work \cite{papernot2018deepknn} proposes a new layer, referred to as the confidence layer, which captures the confidence of the network's predictions. This layer measures the agreement of the predictions of the deep neural network based on the $k$ nearest neighbors to detect nonconformal, out-of-distribution instances. \cite{papernot2018deepknn} highlights the need for interpretable deep learning models, especially in domains where model transparency, explainability, and robustness are critical. Beyond the popularity in CV and NLP,  $k$-NN continues to be a favored approach for classification and regression tasks in tabular and categorical datasets.

\subsection{Anomaly Detection}
Anomaly detection, also known as outlier detection, is a field in data analysis and machine learning that focuses on identifying instances that deviate significantly from the expected behavior of a given dataset. In recent years, the utilization of anomaly detection techniques has expanded across a diverse range of domains. These methods have found application in detecting fraudulent activities within credit cards, insurance, and healthcare sectors, as well as identifying intrusions in cyber-security, pinpointing faults in safety-critical systems, etc. Some of the well known methods for anomaly detection for tabular data include \cite{liu2008isolation, li2022ecod, cbof_he2003discovering, ocsvmli2003improving, breunig2000lof} and \cite{ruff2018deep_svdd}. Traditional methods like Isolation Forests randomly select features and split points to recursively partition data points into subsets, marking path lengths that effectively 'isolate' a datapoint as anomalies. This approach is an ensemble method. Isolation Forests assume that anomalies are sparse and can be isolated easily. If the dataset contains clusters of anomalies that are not well-separated from normal instances, the algorithm's performance may suffer. There are probabilistic methods such as ECOD (Unsupervised Outlier Detection Using Empirical Cumulative Distribution Functions) uses empirical distribution functions to sort the data points and assigns probabilities based on their order. CBLOF (Clustering-Based Local Outlier Factor) assesses the local density deviation of a data point with respect to its neighbors to identify outliers and is a proximity based outlier detection method. While ECOD performs well with unimodal distribution data, it is sensitive to feature dimension noise and may encounter difficulties in accurately identifying outliers in datasets with multiple modes. Additionally, it may face challenges in high-dimensional scenarios, as it focuses on one-dimensional projections.
CBLOF on the other hand focuses on local clusters, potentially missing the global context of the dataset. It necessitates a priori specification of the cluster number, posing a challenge. Unevenly sized clusters may impede its ability to distinguish between normal and outlier instances, particularly within smaller clusters. 
Recently there have been deep learning based approaches such as DeepSVDD which projects high-dimensional data into a latent space using an autoencoder architecture. Anomalies are detected by measuring the distance of data points to the center of a constructed hypersphere in the latent space, employing a threshold to flag anomalous cases based on this distance.
These methods relying on training deep neural networks may exhibit sensitivity to noisy labels when mislabeled instances are present in the training data. This sensitivity can have adverse effects on the model's generalization and its accuracy in detecting outliers. Additionally, these approaches often lack interpretability, making it challenging to understand the rationale behind classifying certain instances as outliers. Furthermore, their performance is often contingent on the careful tuning of hyperparameters within the neural network.

\section{Methods}
\label{sec:methods}

In this section, we introduce the methods through which we enhance $k$-NN, incorporating a novel distance measure and a feature weighting approach, enabling the utilization of innovative techniques and contributing to enhanced performance.  Through the application of these methods to an instance-based $k$-NN base, we leverage the inherent interpretability of the architecture, augmenting it with the following methods and concepts.  For instance, using the formulae that are to be provided, each decision that a model makes can be traced to the individual cases that influenced it. Additionally, the concepts that are introduced are human oriented in terms of both the simplicity of the math involved and the relationship of the various measures to the model itself.

\subsection{Distance Metric}
\label{ssec:dist_metric}
Like many other approaches to $k$-NN, we use Minkowski distance as a starting point

\begin{equation}~\label{eq:minkowski}
d_p(x, y) = \sqrt[\leftroot{-3}\uproot{3}p]{
  \sum_{i \in \Xi} w_i |x_i - y_i|^p
},
\end{equation}

where $p$ is the parameter for the Lebesgue space, $\Xi$ is the feature set, and $w_i$ is the weight for each feature. One problem 
with this distance metric, however, is that distinguishing points becomes more and more difficult in higher dimensions.  
One proposed solution is to use a fractional norm heading towards zero to enable points to be distinguished more easily in high 
dimensional space~\cite{aggarwal2001surprising}. Motivated by this, we derived the Minkowski distance as $p \rightarrow 0$ expressed 
over the feature set $\Xi$

\begin{equation}~\label{eq:minkowski_0}
\lim_{p \rightarrow 0} d_p(x, y) = \sqrt[\leftroot{-3}\uproot{3}|\Xi|]{
  \prod_{i \in \Xi} |x_i - y_i|^{w_i}
},\\
\end{equation}

assuming that the weights $w$ sum to 1.

The above is a geometric mean, which has the useful property of being scale invariant. This derivation presents a problem, however. If 
any of the differences are zero, the entire distance metric will become zero. In order to solve this problem, we use the 
\textit{Łukaszyk–Karmowski} metric as a distance term rather than absolute error. Given two random variables $X$ and $Y$ with 
probability density functions $f(x)$ and $g(x)$ respectively, the LK metric is defined as

\begin{equation}~\label{eq:lk}
d_{LK}(X, Y) = \int_{-\infty}^\infty \int_{-\infty}^\infty |x-y| f(x) g(y) \, dx\, dy .
\end{equation}

We assume that if both points (say $x$ and $y$) are near enough to be worth determining the distance between them, then the distributions and parameters 
for the probability density functions should represent the local data.  The two simple maximum entropy distributions on $(-\infty, 
\infty)$ given a point and a distance around the point are the Laplace distribution (double exponential), where the distance is 
represented as mean absolute error (MAE), and the Gaussian distribution (normal), where the distance is represented as the standard deviation.
We choose the Laplace distribution and derive a closed-form solution of equation \ref{eq:lk} for it. Letting 
$\mu \equiv |x - y|$ and with $b$ being the expected deviation,

\begin{equation}~\label{eq:lk_laplace}
d_{LK}(\mu, b) = \mu + \frac{1}{2} e^\frac{-\mu}{b} \left( 3 b + \mu \right),
\end{equation}
the full derivation of which can be found in the appendix.

In order to employ this measure for our method, we need a value for $b$.  Measurement error may not always be readily 
available, and it does not take into account the additional error among the relationships within the model.  Hence, residuals are 
calculated for each prediction.  The MAE is be calculated for each observation using a leave-one-out approach, where 
instances are removed from the model and each of the held out instance's features are predicted using the rest of the data. (The idea 
is to quantify the uncertainty in the model). These errors can be locally aggregated or can be aggregated across the entire model to 
obtain the expected residual, $r$, for predicting each feature, $i$, as $r_i$. This results in a Minkowski distance metric which uses 
the derived distance term of

\begin{equation}~\label{eq:minkowski_0_deviation}
  \begin{cases}
    \displaystyle
    d_0(x, y) = \sqrt[\leftroot{-3}\uproot{3}|\Xi|]{
      \prod_{i \in \Xi} d_{LK}(|x_i - y_i|, r_i)^{w_i}
    }, & \text{for } p=0\\

    \displaystyle
    d_p(x, y) = \sqrt[\leftroot{-3}\uproot{3}p]{
      \sum_{i \in \Xi} w_i d_{LK}(|x_i - y_i|, r_i)^{p}
    }, & \text{otherwise}.\\

  \end{cases}
\end{equation}

We have found that using the residuals in the $k$-NN system with the above distance metric, calculating new residuals, and then 
feeding these back in, generally yields convergence of the residual values with notable convergence after only 3 or 4 iterations.  
Measuring a distance value for each feature further enables parameterization regarding the type of data a feature holds. For example, 
nominal data can result in a distance of 1 if the values are not equal and 0 if they are equal.  Thus, one-hot encoding, the expansion 
of nominal values into multiple features, is not needed.  Ordinal data can use a distance of 1 between each ordinal type.

\subsection{Inverse Residual Weighting}

Having established a distance metric, we can determine distances between points, but the units of measurement and scales of each feature may be entirely different.  We propose Inverse Residual Weighting (IRW), a maximum entropy method of transforming these each feature difference into surprisal space so that the entire distance itself becomes expected surprisal.

Using our assumption that absolute prediction residuals follow the exponential distribution where the mean value is the feature residual, we can describe the probability a single prediction residual, $v$, being within the feature residual, $r$ as

\begin{equation}
P(v \leq r) = 1 - e^{-\frac{1}{r} v}.
\end{equation}

It then follows that the probability of the prediction residual being outside the feature residual is

\begin{equation}
P(v > r) = e^{-\frac{v}{r} },
\end{equation}

and the surprisal of observing the prediction residual larger than the feature residual is

\begin{equation}
I(v > r) = -ln( P(v > r) ),
\end{equation}

expanded as

\begin{align*}
I(v > r) = -ln( e^{-\frac{v}{r} } )
&= \frac{v}{r}.
\end{align*}

In light of this, we observe that to find the surprisal of an observed residual, we can simply divide by the feature residual. This is the motivation for using IRW, where the inverse of feature residuals are used for feature weights when computing distances.

As previously described, we are able to compute a residual for each feature as the mean absolute deviation between the observed
values and predicted values for the feature. We can express the feature residual $r_i$ as

\begin{equation}~\label{eq:feature_residual}
 r_i = \frac{1}{|X|} \sum_{x_j \in X} | x_{i,j} - \hat{x}_{i,j} |,
\end{equation}

where $x_{i,j}$ represents the $i^{th}$ feature value of case $x_j$ and $\hat{x}_{i,j}$ represents the prediction for that specific value. Then this feature residual can be used to determine feature weights,
$w_i$, which can then be expressed as

\begin{equation}~\label{eq:inverse_residual_feature_weight}
 w_i = \frac{1}{r_i^p}.
\end{equation}

Using the inverse of the residual as the weight for each feature allows the distance contributed by each feature to
be in the same space as one another.  This gives the distance function scale invariance across varying feature types and scales, which 
solves one common challenge of using nearest neighbors approaches.  Additionally, using IRW allows the model to emphasize features with strong 
relationships and reduce the influence features that appear to be significantly noisy or generally unpredictable.
For models with a designated target feature, feature weights can further augmented using Mean Decrease in Accuracy (MDA) or similar 
techniques that attempt to capture the predictive power of a feature. Additionally, we are actively researching methods of incorporating MDA techniques alongside IRW in targetless applications of our methods.

Furthermore, scaling by the inverse residual feature weights enables the system to interpret distances in surprisal space. Being in surprisal space allows us to utilize a maximum entropy assumption and the Laplace distribution to measure observed residuals in terms of surprisal. These surprisal values can then be utilized for various metrics and downstream tasks. Specifically we use these surprisal values to compute surprisal ratios that we refer to as convictions, which is covered in detail in the concepts section.

\section{Concepts}
\label{sec:concepts}
In this section, we introduce human-oriented concepts which enable or enhance interpretable analyses and applications of the above methods to common tasks including classification, regression, feature selection, and anomaly detection. Many of these concepts are 
naturally understandable, being ratios. Additionally, they provide insight that lends itself naturally to strong performance on many 
difficult machine learning tasks.

\subsection{Distance Contribution}

The distance contribution reflects how much distance a point contributes to a graph connecting the nearest neighbors, which is the 
inverse of the density of points over a unit of distance in the Lebesgue space.  The harmonic mean of the distance contribution 
reflects the inverse of the inverse distance weighting often employed with $k$-NN, though other techniques may be substituted if inverse 
distance weighting is not employed. We define the distance contribution as:

\begin{equation}~\label{eq:distance_contribution}
  \phi(x) = \bigg( \frac{1}{k} \sum_{x_j \in X_k} \frac{1}{d(x, x_j)} \bigg) ^{-1},
\end{equation}

where $X_k$ is the set of nearest neighbors to point $x$ and $d$ is the distance function.  This is a harmonic mean over 
the distances to each nearest neighbor.  Note that the properties of the previously defined distance metric are useful here to 
prevent divisions by zero.

We can quantify the information needed to express a distance contribution $\phi(x)$ by transforming it into a probability.  
We begin by selecting the exponential distribution to describe the distribution of residuals as it is the maximum entropy 
distribution constrained by the first moment.  We represent this in typical nomenclature for the exponential distribution 
using $p$ norms.

\begin{equation}
  \frac{1}{\lambda} = \|r(x)\|_p.
\end{equation}

We can directly compare the distance contribution and p-normed magnitude of the residual.  This is because the 
distance contribution and the norm of the residual are both on the same scale, with the distance contribution 
being the expected distance of new information that the point adds to the model, and the norm of the residual is 
the expected distance of deviation.  Given the entropy maximizing assumption of the exponential distribution of the 
distances, we can then determine the probability that a distance contribution is greater than or equal to the 
magnitude of the residual $\|r(x)\|_p$ in the form of cumulative residual entropy~\cite{Rao_2004} as

\begin{equation}
  P(\phi(x) \geq \|r(x)\|_p) = e^{-\frac{1}{\|r(x)\|_p} \cdot \phi(x)}.
\end{equation}
We then convert the probability to self-information as
\begin{equation}
  I(x) = -\ln P(\phi(x) \geq \|r(x)\|_p),
\end{equation}
which simplifies to
\begin{equation}~\label{eq:prediction_conviction_self_info}
  I(x) = \frac{\phi(x)}{\|r(x)\|_p}.
\end{equation}

\subsection{Conviction}

If we have some form of prior distribution of data given all of the information observed up to that point, 
the surprisal is the amount of information gained when we observe a new sample, event, case, or state change 
and update the prior distribution to form a new posterior distribution after the event.  The surprisal of an 
event of observing a random variable $x \sim X$ is defined as $I(x) = -\ln p(x)$.  Thus, the conviction, 
$\pi$, can be expressed as
\begin{equation}~\label{eq:conviction}
	\pi(x) = \frac{\mathbb{E}[I(X)]}{I(x)}.
\end{equation}

By computing this ratio for different types of information, we derive several different types of 
\textit{conviction} with different uses in various applications: \textit{familiarity conviction}, 
\textit{similarity conviction}, and \textit{residual conviction}.

\subsubsection{Familiarity Conviction}

\emph{Familiarity conviction} is a metric for describing surprisal of points in a model relative to the 
training data.  Consider a data set that has data points at regular intervals, such as a data point for each 
corner in a grid.  Now consider a new point is added that is very close to one of the existing corner points.
This new point should be quite easy to predict as it is close to an existing point, making it unsurprising.
However, given this grid data, familiarity conviction would indicate a higher surprisal for such a 
point even though it is easy to label because the point is unusual with regard to the even distribution of 
the rest of the data points.  This new point does not form another corner of the grid.  These properties make 
familiarity conviction valuable for sanitizing data and reducing data as well as extracting patterns and 
anomalies, as is discussed in other sections.

Familiarity conviction is based on the distance metric described previously.  As long as a low or zero value of 
$p$ is used in $L_p$ space metrics for similarity, familiarity conviction is independent of the scale of the data provided and does not overreact to feature dominance based on feature scale and range. Given a set of points 
$X \subset \mathbb{R}^z, \ \forall x \in \mathbb{X}$ and an integer $1 \leq k < |X|$ we define the 
distance contribution probability distribution, $C$ of $X$ to be the set
\begin{equation}~\label{eq:distance_probs}
  C = \left\{
    \frac{\phi(x_1)}{\sum_{j=1}^{n}\phi(x_j)},
    \frac{\phi(x_2)}{\sum_{j=1}^{n}\phi(x_j)},
    \dots,
    \frac{\phi(x_n)}{\sum_{j=1}^{n}\phi(x_j)}
  \right\}, 
\end{equation}
for a function $\phi:X \to \mathbb{R}$ that returns the distance contribution. Note that because $\phi(0) = 
\infty$ may be true under some circumstances, multiple identical points may need special consideration, such 
as splitting the distance contribution among those points. Clearly $C$ is a valid probability distribution.  
We will use this fact to compute the amount of information in $C$.  The point probability of a point $x_{i}, 
j=1, 2, \dots, n$ is
\begin{equation}
  l(i) = \frac{\phi(x_i)}{\sum_{j=1}^{n}\phi(x_j)},
  \label{eq:pprob}
\end{equation}
where we see the index $j$ is assigned the probability of the indexed point's distance contribution.

We assume the set of random variables that characterize the discrete distribution of point probabilities, $L$, 
is the set of $L = \{l(1), l(2), \dots, l(n)\}$. Because we have no additional knowledge of the distribution of 
points other than they follow the distribution of the data, we assume $L$ is uniform as the distance probabilities 
have no trend or correlation. Then, the familiarity conviction of a point $x_j\in X$ is defined as

\begin{equation}~\label{eq:familiarity_conviction}
  \pi_f(x_j) = \frac{ 
    \frac{1}{|X|} \sum\limits_{j} D_{KL}\left(L || L - \{j\}\cup \mathbb{E} l(j) \right)
  }{ 
    D_{KL}\left(L || L - \{x_j\}\cup \mathbb{E} l(j) \right) 
  },
\end{equation}

where $D_{KL}$ is the Kullback-Leibler divergence. Since we assume $L$ is uniform, we have that the expected probability $\mathbb{E} l(j) = \frac{1}{n}$.

Familiarity conviction is well suited for anomaly detection, particularly at detecting inliers, which would have familiarity 
conviction significantly smaller than 1. This performance comes at the cost of computational complexity.

\subsubsection{Similarity Conviction}

\emph{Similarity Conviction} is another method to evaluate the surprisal of a point in the data relative to the distribution
of data that make up the point's nearest neighbors. Similarity conviction is defined as the expected distance contribution of 
the point divided by the point's observed distance contribution.  To get the expected distance contribution of a point, the 
distance contributions of its nearest $k$ neighbors are computed and then averaged.  Using the local model 
of the point to get an expected distance contribution gives us a measure of conviction that leverages the contextual information 
about the sparsity in the local model.

Similarity conviction can be used as a tool to identify anomalies in the data, whether looking for inliers or outliers.  Inliers 
will have uncharacteristically low distance contribution, and consequently have high values for similarity conviction.  Similarly, 
outliers should have higher distance contributions than their local model which gives them low values for similarity conviction.  
Non-anomalous data should be expected to have similarity conviction values around 1.0, since the expected distance contribution is 
to be expected.  Similarity conviction is less computationally expensive than familiarity conviction, but may not perform as well 
at identifying certain inliers.

Similarity conviction, $\pi_s$, can be expressed as:
\begin{equation}~\label{eq:similarity_conviction}
	\pi_s(x) = \frac{\mathbb{E}[\phi(x)]}{\phi(x)}.
\end{equation}

Using the average distance contribution of the local model as the expected distance contribution, $\mathbb{E}[\phi(x)]$ can be expressed by:

\begin{equation}~\label{eq:expected_distance_contribution}
 \mathbb{E}[\phi(x)] = \bigg(  \frac{1}{k}    \sum_{x_j \in X_k} \phi(x_j) \bigg).
\end{equation}

\subsubsection{Residual Conviction}

Examining \emph{residual conviction} provides insight into the model's uncertainty for a feature prediction. Residual conviction 
is calculated as the expected model residual for a feature divided by the computed prediction residual for that feature. The 
expected model residual is calculated by for taking the mean of the residuals in the local model of its nearest $k$ neighbors 
around the predicted feature, thus the residual conviction for feature $i$ of point $x$ is

\begin{equation}~\label{eq:residual_conviction}
  \pi_r(x, \hat{x}, i) = \frac{\displaystyle
    \frac{1}{k} \sum_{x_i \in X_k} |x_i - \hat{x}_{i}|
    }{
      |x - \hat{x}|
  },\\
\end{equation}
  
where $X_k$ is the set of points in the local model around point $x$. This ratio quantifies the difficulty of individual case's feature prediction, with prediction certainty decreasing as the conviction approaches 0.  
In more practical terms, residual conviction serves to characterize how uncertain one or more predictions are 
relative to how uncertain they are expected to be.  This can be used to explain model decisions.  If a decision 
is incorrect but has a residual conviction $\approx 1$, then this uncertainty is likely due to uncertainty in 
the data rather than the model.

\section{Applications}
\label{sec:applications}
% Experimental results go here.
In this section we demonstrate the performance of the above methods and concepts on various machine 
learning tasks.  Namely, classification, regression, and anomaly detection. In general, we see that $k$-NN using these 
enhancements consistently performs near or above state of the art while maintaining strong interpretability and flexibility. 
%\footnote{Code will be made publicly available upon acceptance.}. 

\subsection{Classification and Regression}
We conducted a comprehensive series of experimental comparisons on a diverse set of algorithms. We first perform classification and regression across 308 PMLB datasets \cite{romano2021pmlb}. (146 for classification and  162 for regression) \footnote{Kindly refer to the appendix for more information on the datasets used and for additional experiments and details.}. and compare our approach across gradient boosted trees, traditional $k$-nearest neighbors, logistic regression (for classification), regularized least squares (for regression), neural networks, random forests and Light-GBM. A stratified sampling of data having cells $N \cdot d \leq 400000$ was chosen. 
To ensure robustness and reliability, each classification and regression experiment was iterated 30 times with varying random seeds, and the resulting metric averages were computed for statistical significance. For classification tasks, we present mean, precision, recall, and Matthews Correlation Coefficient (MCC) as evaluation metrics. In regression, Mean $R^2$, mean absolute error (MAE), mean square error (MSE) and Spearman coefficient were calculated. The consolidated results are detailed in Table \ref{tab:classification} and Table \ref{tab:regression}, providing a comparative perspective against a diverse set of algorithms. It is worth noting that our proposed method consistently outperforms all other classification algorithms in terms of accuracy and precision, while also demonstrating competitive results in regression.

% \subsection{}

\begin{table*}[ht!]
\centering
\scriptsize{}
\caption{Classification Results across 146 PMLB Datasets}
\text{(\color{blue}{\textbf{Blue}} values indicate the best performance; \color{brown}{\textbf{Brown}} values indicate the second-best performance )}
\begin{center}
\vskip 0.15in
\begin{small}
\begin{sc}
\begin{tabular}{lrrrrrrrr}
\toprule
{Classification} &      \textbf{Ours} &        GB &       KNN &        LR &        NN &        RF &         LGBM \\
\midrule
Mean Accuracy $(\%)$ $\uparrow$ &  \color{blue}{\textbf{82.2278}} &  \color{brown}{\textbf{81.9668}} &  79.3017 &    79.0799 &  79.7512 &  81.4218  & 81.9154\\
Mean Precision $\uparrow$ &  \color{blue}{\textbf{0.786554}} &  0.774766 &  0.746573 &    0.743946 &  0.732757 &  0.772808 &  \color{brown}{\textbf{0.782817}}\\
Mean Recall $\uparrow$    &  \color{brown}{\textbf{0.770464}} &  0.764163 &  0.719889 &    0.731404 &  0.736894 &  0.756036 & \color{blue}{\textbf{0.779027}}\\
Mean MCC $\uparrow$       & \color{brown}{\textbf{0.644526}} &  0.628351 &  0.562876 &    0.575105 &  0.584838 &  0.618480 &  \color{blue}{\textbf{0.653397}}\\
\bottomrule
\end{tabular}
\label{tab:classification}
\end{sc}
\end{small}
\end{center}
\vskip -0.1in
\end{table*}

\begin{table*}[ht!]
\centering
\scriptsize{}
\caption{Regression Results across 162 PMLB Datasets}
\text{(\color{blue}{\textbf{Blue}} values indicate the best performance; \color{brown}{\textbf{Brown}} values indicate the second-best performance )}
\begin{center}
\vskip 0.15in
\begin{small}
\begin{sc}
\begin{tabular}{lrrrrrrrr}
\toprule
{Regression} &       \textbf{Ours} &         GB &        KNN &         Linear &          NN &         RF &      LGBM \\
\midrule
$R^2$ Mean  $\uparrow$    &   \color{brown}{\textbf{0.857244}} &   \color{blue}{\textbf{0.864342}} &   0.724857 &    0.509989 &   0.727337 &   0.855368 &  0.818680  \\
MAE   $\downarrow$         &  \color{brown}{\textbf{0.841702}} &   0.851328 &   1.038063 &    1.975748 &   1.008824 &   \color{blue}{\textbf{0.827424}} &  1.069768  \\
MSE  $\downarrow$           & \color{brown}{\textbf{10.168815}} &  10.459248 &  12.222883 &   36.569369 &  11.598598 &  \color{blue}{\textbf{9.640429}} &  20.453173  \\
Spearman coeff. $\uparrow$ &  0.916272 &   \color{blue}{\textbf{0.925263}} &   0.832777 &    0.719865 &   0.821594 &   \color{brown}{\textbf{0.917626}} &  0.913498 \\
\bottomrule
\end{tabular}
\end{sc}
\end{small}
\end{center}
\label{tab:regression}
\end{table*}

\subsection{Anomaly Detection}
Using the defined conviction metrics, we can judge whether or not a data point is an anomaly on a standardized scale.
To evaluate the the accuracy of this method, we present results on anomaly detection on 20 datasets from Outlier Detection Datasets (ODDS) \footnote{Kindly refer to the appendix for dataset related details.} \cite{Rayana:2016}.
These datasets have ground truth labels indicating which data points are anomalous, which makes them ideal for this analysis.
We utilize the previously established method of evaluating the conviction values of each point and compare to the results of using many of the popular anomaly detection methods as shown in Table \ref{tab:anomaly_detection_results_average}.
Specifically, we trained our model by splitting our dataset into two parts (train and test). The training set comprised solely of inliers, and a test set encompassing both inliers and outliers, with a notable prevalence of inliers. Since the ODDS dataset has ground truth labels for both inliers and outliers, 
% with half of the dataset's non-anomalous data, then performed anomaly detection
% on the remaining half of the non-anomalies and all of the anomalies of the data. Then using the predictions and
we used the ground truth labels to compute F1 scores to measure the performance of the anomaly detection benchmark routine.
For our methods, we simply computed the conviction (similarity conviction or familiarity conviction) and compared it to a threshold of 0.7. If the conviction fell below the threshold, then it was classified as an anomaly. In practice we would recommend tuning this threshold per dataset, but here we show that picking a conviction level of 0.7 for all datasets (wihout choosing it in a dataset specific manner), our method achieves the highest $F1$ scores in 12 of the 20 datasets, surpassing the performance of all other outlier detection methods.

In Table~\ref{tab:anomaly_detection_results_average}, we show the average F1 score for each method across the
20 ODDS datasets. To see the results per dataset, please refer to the appendix.

\begin{table*}[ht!]
  \centering
   \scriptsize{}
    \caption{Mean F1 scores for Anomaly detection across 20 ODDS datasets}
    \text{(\textbf{Bold} values indicate the best performance)}
\vskip 0.15in
\begin{center}
\begin{small}
\begin{sc}
\begin{tabular}{lc}
\toprule
Method &   Mean F1 Score $\uparrow$ \\
\midrule
\textbf{Ours (Familiarity Conviction) }&            0.32 \\
\textbf{Ours (Similarity Conviction)}  &            \textbf{0.49} \\
One Class SVM  \cite{ocsvmli2003improving}                &            0.22 \\
Isolation Forest \cite{liu2008isolation}              &            0.38 \\
CBLOF     \cite{cbof_he2003discovering}                    &            0.37 \\
Local Outlier Factor   \cite{breunig2000lof}         &            0.19 \\
ECOD     \cite{li2022ecod}                      &            0.32 \\
DeepSVDD  \cite{ruff2018deep_svdd}                    &            0.45 \\
\bottomrule
\end{tabular}
\end{sc}
\end{small}
\end{center}
\vskip -0.1in
  \label{tab:anomaly_detection_results_average}
\end{table*}

It is worth noting that certain methodologies, such as  CBLOF (Clustering-Based Local Outlier Factor), LOF (Local Outlier Factor), and ECOD (Extended Connectivity-Based Outlier Detection) usually incorporate a distinct partition exclusively composed of inliers during the training phase. Though this is not necessary, these methods can benefit from inlier-based training partitions. In contrast, our approach which harnesses the notion of familiarity conviction, allows us the capability to identify anomalies without necessitating an explicit `inlier' dataset. This innovation enables us to gauge the uncertainty inherent in our model and promptly identify anomalous instances in a real-time manner. 

\begin{figure}[ht!]
    \centering
    \includegraphics[width=0.99\linewidth]{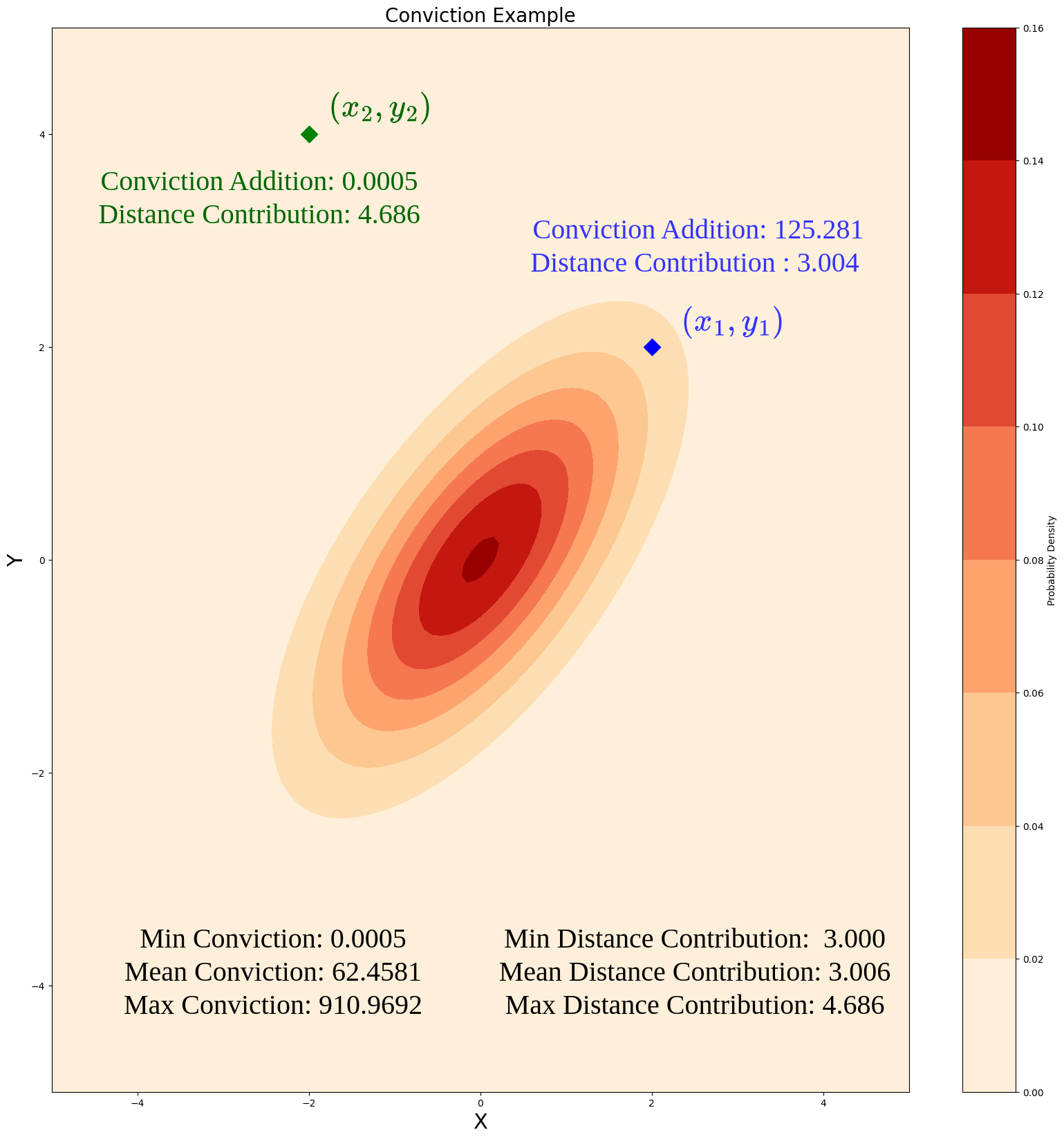}
    \caption{Depiction of familiarity conviction addition and distance contribution of two given points $(x_1, y_1)$ and $(x_2, y_2)$ given data distributed normally.}
    \label{fig:conviction_eg}
\end{figure}

We demonstrate this on a toy dataset as shown in Figure \ref{fig:conviction_eg}. 
 Consider a toy training dataset $D$ in Figure \ref{fig:conviction_eg} which is sampled from  a random variable $X \sim \mathcal{N}(0, \Sigma^2)$ and we observe two points $(x_1, y_1)$ (blue) and $(x_2, y_2)$ (green) within the data.  Given this data, familiarity conviction allows us to measure how close a point is to existing data.  Using our notion of surprisal, we can observe that $(x_2, y_2)$ has high surprisal and therefore low conviction. Moreover, the distance contribution is higher than the mean distance contribution of the entire dataset. This allows for us to detect it as anomalous without having requirement of a separate inlier partition.

\section{Limitations and Future Work}
The present methodology, while promising, exhibits certain limitations in terms of the scale of data it can effectively handle. As an extension of the foundational $k$-Nearest Neighbors framework, this approach necessitates that the data used for model fitting is constrained to the memory capacity of the a machine. 
Secondly, the challenge lies in determining the conviction threshold beforehand, as it depends on factors such as the contamination level within the anomalous dataset or the absence of an inlier training set (for similarity conviction). Furthermore, placing additional emphasis on interpretability may introduce a trade-off in classification and regression performance, as the model becomes less reliant on spurious correlations within the data.
To address the issue of scale we have implemented techniques that make querying the dataset more efficient than many standard methods. Specifically, in practice we use an efficient branch-and-bound implementation that combines efficient use of bit vectors to reduce the total compute required for all but the most pathological datasets. Furthermore, we are looking into sampling strategies of best representing the variance of our original data as well as data ablation techniques which could allow us to store more information in less space by intelligently adjusting weights of trained cases. We have also begun to probe the robustness of our method against inference-time adversarial attacks. Owing to the lack of gradient-based optimization in our approach and the adeptness at outlier detection tasks, early results have shown significant promise of our method's robustness to out of distribution data across tabular as well as image datasets. Although this facet remains part of our future work, early indications of its resilience against such challenges substantiate the potential for our approach to thrive in safety-critical contexts.

\section{Discussion and Conclusion}
\label{sec:conclusion}

In summary, we propose several enhancements to the traditional $k$-NN algorithm from the perspective of information theory. 
In particular, our method utilizes the Łukaszyk–Karmowski (LK) distance tailored to Laplace distributions, effectively mitigating the problem of zero distances predicated on data uncertainty.  Furthermore, by leveraging Inverse Residual Weighting (IRW), we convert our distance measurements into the realm of surprisal space. Using the notion of surprisal, we define a new concept of conviction with which we are able to compute interpretable measures of the importance and surprisal for each data point. Finally, these enhancements have increased the effectiveness of $k$-NN while maintaining its natural interpretability.
Since our method utilizes nearest-neighbors, it can effectively estimate the underlying density of the data, contributing to its versatility in various statistical and machine learning applications. Unlike traditional methods that rely on post-model interpretability tools, our approach directly addresses data and feature uncertainty. By leveraging the aforementioned tools,  we can determine data point weights as well as feature contributions by calculating the conditional entropy for adding a feature without the need for explicit model training. This also allows us to compute MDA and MAE along with feature contributions. Our method provides detailed data point influence weights with perfect attribution and can be used to query counterfactuals. Moreover, by calculating SHAP over feature sets sampled from the entire feature space, we can obtain a more reliable estimate of SHAP that is robust to multicollinearity and feature order. 
Lastly, we conducted an extensive analysis on 308 datasets for classification and regression, alongside an additional 20 ODDS datasets for anomaly detection. In conclusion, we see that this approach aligns with human understanding of decision-making from data, such as similarity, differences and causality and facilitates a clearer understanding of a model's decision-making process.

\bibliography{bibliography}
\bibliographystyle{icml2024}

%%%%%%%%%%%%%%%%%%%%%%%%%%%%%%%%%%%%%%%%%%%%%%%%%%%%%%%%%%%%%%%%%%%%%%%%%%%%%%%
%%%%%%%%%%%%%%%%%%%%%%%%%%%%%%%%%%%%%%%%%%%%%%%%%%%%%%%%%%%%%%%%%%%%%%%%%%%%%%%
% APPENDIX
%%%%%%%%%%%%%%%%%%%%%%%%%%%%%%%%%%%%%%%%%%%%%%%%%%%%%%%%%%%%%%%%%%%%%%%%%%%%%%%
%%%%%%%%%%%%%%%%%%%%%%%%%%%%%%%%%%%%%%%%%%%%%%%%%%%%%%%%%%%%%%%%%%%%%%%%%%%%%%%
\newpage
\appendix
\onecolumn

% You can have as much text here as you want. The main body must be at most $8$ pages long.
% For the final version, one more page can be added.
% If you want, you can use an appendix like this one.  

% The $\mathtt{\backslash onecolumn}$ command above can be kept in place if you prefer a one-column appendix, or can be removed if you prefer a two-column appendix.  Apart from this possible change, the style (font size, spacing, margins, page numbering, etc.) should be kept the same as the main body.
\section{Derivation of Łukaszyk–Karmowski (LK) with Laplace Distributions}
To prove Equation 4 of our work, we begin with the expected distance between two random variables $X$ and $Y$ given two probability density functions, $f(x)$ and $g(y)$ as
\begin{equation}
d(X, Y) = \int_{-\infty}^\infty \int_{-\infty}^\infty |x-y| f(x) g(y) \, dx\, dy.
\end{equation}

Using two Laplace distributions with means $\mu_1$ and $\mu_2$ and expected distance from the mean $b_1$ and $b_2$, we can express the probability density functions as
\begin{equation}
f(x) = \frac{1}{2 b_1} e^{\frac{-|x - \mu_1|}{b_1}}
\end{equation}
and 
\begin{equation}
g(y) = \frac{1}{2 b_2} e^{\frac{-|y - \mu_2|}{b_2}}
\end{equation}
respectively.

Substituting in the Laplace distributions into the expected distance, we can simplify this slightly as
\begin{align*}
d(X, Y) &= \int_{-\infty}^\infty \int_{-\infty}^\infty |x-y| \cdot \frac{1}{2 b_1} e^{\frac{-|x - \mu_1|}{b_1}} \cdot \frac{1}{2 b_2} e^{\frac{-|y - \mu_2|}{b_2}} \, dx\, dy \\
&= \frac{1}{4 b_1 b_2} \int_{-\infty}^\infty \int_{-\infty}^\infty |x-y| \cdot e^{\frac{-|x - \mu_1|}{b_1}} \cdot e^{\frac{-|y - \mu_2|}{b_2}} \, dx\, dy.
\end{align*}

We further assume that $b_1 = b_2$, and use $b$ in place of both, which assumes that the error is the same throughout the space and simplify further as
\begin{equation}
\label{eq:laplace_lk_symmetric_b}
% d_{v \leq \mu_1}(X, Y) = \frac{1}{4 b^2} \int_{-\infty}^{\infty} \int_{-\infty}^{\infty} |x-y| \cdot e^{\frac{-|\mu_1 - x|}{b}} \cdot e^{\frac{-|\mu_2 - y|}{b}} \, dx\, dy.
d(X, Y) = \frac{1}{4 b^2} \int_{-\infty}^{\infty} \int_{-\infty}^{\infty} |x-y| \cdot e^{\frac{-|\mu_1 - x|}{b}} \cdot e^{\frac{-|\mu_2 - y|}{b}} \, dx\, dy.
\end{equation}

Because we only have one value for $b$, we can assume that $\mu_1 \leq \mu_2$ without loss of generality because we can just exchange the values if this is not true, and in the end we will adjust the formula to remove this assumption.  There exist 3 regions of the space for $x$ which are $x \leq \mu_1$, $\mu_1 < x \leq \mu_2$, and $\mu_2 < x$.

\subsection{$x \leq \mu_1, y \leq \mu_1$}

Rewriting Equation~\ref{eq:laplace_lk_symmetric_b} for the part of the space where $x < \mu_1, y < \mu_1$ is
\begin{align*}
d_{x \leq \mu_1, y \leq \mu_1}(X, Y) &= \frac{1}{4 b^2} \int_{-\infty}^{\mu_1} \int_{-\infty}^{\mu_1} |y - x| \cdot e^{\frac{-(\mu_1 - x)}{b}} \cdot e^{\frac{-(\mu_2 - y)}{b}} \, dx\, dy \\
 &= \frac{1}{4 b^2} \int_{-\infty}^{\mu_1} \left( \int_{-\infty}^{\mu_1} |y - x| \cdot e^{\frac{-(\mu_1 - x)}{b}} \, dx \right) \cdot e^{\frac{-(\mu_2 - y)}{b}} \, dy \\
  &= \frac{1}{4 b^2} \int_{-\infty}^{\mu_1} \left( \int_{-\infty}^{y} (y - x) \cdot e^{\frac{-(\mu_1 - x)}{b}} \, dx + \int_{y}^{\mu_1} (x - y) \cdot e^{\frac{-(\mu_1 - x)}{b}} \, dx \right) \cdot e^{\frac{-(\mu_2 - y)}{b}} \, dy \\
  &= \frac{1}{4 b^2} \int_{-\infty}^{\mu_1} \left( b^2 e^\frac{y-\mu_1}{b} + b^2 e^\frac{y-\mu_1}{b} - b y + b \mu_1 - b^2 \right) \cdot e^{\frac{-(\mu_2 - y)}{b}} \, dy \\
  &= \frac{1}{4 b} \int_{-\infty}^{\mu_1} \left( 2 b e^\frac{y-\mu_1}{b} - y + \mu_1 - b \right) \cdot e^{\frac{-(\mu_2 - y)}{b}} \, dy \\
  &= \frac{1}{4 b} b^2 e^\frac{-(\mu_2 - \mu_1)}{b}\\
  &= \frac{1}{4} b e^\frac{-(\mu_2 - \mu_1)}{b}.
\end{align*}

\subsection{$x \leq \mu_1, \mu_1 < y \leq \mu_2$}

Rewriting Equation~\ref{eq:laplace_lk_symmetric_b} for the part of the space where $x < \mu_1, \mu_1 < y < \mu_2$ is
\begin{align*}
d_{x \leq \mu_1, \mu_1 < y \leq \mu_2}(X, Y) &= \frac{1}{4 b^2} \int_{\mu_1}^{\mu_2} \int_{-\infty}^{\mu_1} (y - x) \cdot e^{\frac{-(\mu_1 - x)}{b}} \cdot e^{\frac{-(\mu_2 - y)}{b}} \, dx\, dy \\
&= \frac{1}{4 b^2} \int_{\mu_1}^{\mu_2} \left( b y - b \mu_1 + b^2\right) \cdot e^{\frac{-(\mu_2 - y)}{b}} \, dy \\
&= \frac{1}{4 b} \int_{\mu_1}^{\mu_2} \left( y - \mu_1 + b\right) \cdot e^{\frac{-(\mu_2 - y)}{b}} \, dy \\
&= \frac{1}{4 b} \left( b \mu_2 - b \mu_1 \right) \\
&= \frac{1}{4} \left( \mu_2 - \mu_1 \right) .
\end{align*}

\subsection{$x \leq \mu_1, \mu_2 < y$}

Rewriting Equation~\ref{eq:laplace_lk_symmetric_b} for the part of the space where $x < \mu_1, \mu_2 < y$ is
\begin{align*}
d_{x \leq \mu_1, \mu_2 < y}(X, Y) &= \frac{1}{4 b^2} \int_{\mu_2}^{\infty} \int_{-\infty}^{\mu_1} (y - x) \cdot e^{\frac{-(\mu_1 - x)}{b}} \cdot e^{\frac{-(y - \mu_2)}{b}} \, dx\, dy\\
&= \frac{1}{4 b^2} \int_{\mu_2}^{\infty} \left( \int_{-\infty}^{\mu_1} (y - x) \cdot e^{\frac{-(\mu_1 - x)}{b}} \, dx\, \right) \cdot e^{\frac{-(y - \mu_2)}{b}} dy\\
&= \frac{1}{4 b^2} \int_{\mu_2}^{\infty} \left( b y - b \mu_1 + b^2  \right) \cdot e^{\frac{-(y - \mu_2)}{b}} dy\\
&= \frac{1}{4 b} \int_{\mu_2}^{\infty} \left( y - \mu_1 + b  \right) \cdot e^{\frac{-(y - \mu_2)}{b}} dy\\
 &= \frac{1}{4b} \left( b \mu_2 - b \mu_1 + 2 b^2 \right) \\
 &= \frac{1}{4} \left( \mu_2 - \mu_1 + 2 b \right).
\end{align*}

\subsection{$\mu_1 < x \leq \mu_2, y \leq \mu_1$}

Rewriting Equation~\ref{eq:laplace_lk_symmetric_b} for the part of the space where $\mu_1 < x < \mu_2, y < \mu_1$ is
\begin{align*}
d_{\mu_1 < x \leq \mu_2, y \leq \mu_1}(X, Y) &= \frac{1}{4 b^2} \int_{-\infty}^{\mu_1} \int_{\mu_1}^{\mu_2} (x - y) \cdot e^{\frac{-(x - \mu_1)}{b}} \cdot e^{\frac{-(\mu_2 - y)}{b}} \, dx\, dy\\
 &= \frac{1}{4 b^2} \int_{-\infty}^{\mu_1} \left( \int_{\mu_1}^{\mu_2} (x - y) \cdot e^{\frac{-(x - \mu_1)}{b}} \, dx \right) \cdot e^{\frac{-(\mu_2 - y)}{b}} \, dy\\
 &= \frac{1}{4 b^2} \int_{-\infty}^{\mu_1} \left( e^\frac{\mu_1 - \mu_2}{b} \left( b y - b \mu_2 - b^2 \right) - b y + b \mu_1 + b^2 \right) \cdot e^{\frac{-(\mu_2 - y)}{b}} \, dy\\
 &= \frac{1}{4 b} \int_{-\infty}^{\mu_1} \left( e^\frac{\mu_1 - \mu_2}{b} \left( y - \mu_2 - b \right) - y + \mu_1 + b \right) \cdot e^{\frac{-(\mu_2 - y)}{b}} \, dy\\
 &= \frac{1}{4 b} \left( 2 b^2 e^\frac{\mu_1-\mu_2}{b} + e^\frac{2 \mu_1 - 2 \mu_2}{b} \left(-b \mu_2 + b \mu_1 - 2 b^2 \right) \right) \\
 &= \frac{1}{4} \left( 2 b e^\frac{\mu_1-\mu_2}{b} + e^\frac{2 \mu_1 - 2 \mu_2}{b} \left( \mu_1 - \mu_2 - 2 b \right) \right). 
\end{align*}

\subsection{$\mu_1 < x \leq \mu_2, \mu_1 < y \leq \mu_2$}

Rewriting Equation~\ref{eq:laplace_lk_symmetric_b} for the part of the space where $\mu_1 < x < \mu_2, \mu_1 < y < \mu_2$ is
\begin{align*}
d_{\mu_1 < x \leq \mu_2, \mu_1 < y \leq \mu_2}(X, Y) &= \frac{1}{4 b^2} \int_{\mu_1}^{\mu_2} \int_{\mu_1}^{\mu_2} |y - x| \cdot e^{\frac{-(x - \mu_1)}{b}} \cdot e^{\frac{-(\mu_2 - y)}{b}} \, dx\, dy\\
 &= \frac{1}{4 b^2} \int_{\mu_1}^{\mu_2} \left( \int_{\mu_1}^{\mu_2} |y - x| \cdot e^{\frac{-(x - \mu_1)}{b}} \, dx \right) \cdot e^{\frac{-(\mu_2 - y)}{b}} \, dy\\
 &= \frac{1}{4 b^2} \int_{\mu_1}^{\mu_2} \left( \int_{\mu_1}^{y} (y - x) \cdot e^{\frac{-(x - \mu_1)}{b}} \, dx  + \int_{y}^{\mu_2} (x - y) \cdot e^{\frac{-(x - \mu_1)}{b}} \, dx \right) \cdot e^{\frac{-(\mu_2 - y)}{b}} \, dy\\
 &= \frac{1}{4 b^2} \int_{\mu_1}^{\mu_2} \bigg( \left( b^2 e^\frac{\mu_1 - y}{b} + b y - b \mu_1 - b^2 \right) \\
 &\qquad + \left( b^2 e^\frac{\mu_1 - y}{b} + e^\frac{\mu_1 - \mu_2}{b} \left( b y - b \mu_2 - b^2 \right) \right) \bigg) \cdot e^{\frac{-(\mu_2 - y)}{b}} \, dy\\
 &= \frac{1}{4 b} \int_{\mu_1}^{\mu_2} \left( 2 b e^\frac{\mu_1 - y}{b} + y - \mu_1 - b + e^\frac{\mu_1 - \mu_2}{b} \left( y - \mu_2 - b \right) \right) \cdot e^{\frac{-(\mu_2 - y)}{b}} \, dy\\
 &= \frac{1}{4 b} \bigg( b \mu_2 - b \mu_1 - 2 b^2 + e^\frac{\mu_1 - \mu_2}{b} \left( 2 b \mu_2 - 2 b^2 \right) \\
 &\qquad + e^\frac{\mu_1 - \mu_2}{b} \left( -2 b \mu_1 + 2 b^2\right) + e^\frac{2 \mu_1 - 2 \mu_2}{b} \left( b \mu_2 - b \mu_1 + 2 b^2 \right) \bigg) \\
 &= \frac{1}{4} \left( \mu_2 - \mu_1 - 2 b + e^\frac{\mu_1 - \mu_2}{b} \left( 2 \mu_2 - 2 \mu_1 \right) + e^\frac{2 \mu_1 - 2 \mu_2}{b} \left(\mu_2 - \mu_1 + 2 b \right) \right).
\end{align*}

\subsection{$\mu_1 < x \leq \mu_2, \mu_2 < y$}

Rewriting Equation~\ref{eq:laplace_lk_symmetric_b} for the part of the space where $\mu_1 < x < \mu_2, \mu_2 < y$ is
\begin{align*}
d_{\mu_1 < x \leq \mu_2, \mu_2 < y}(X, Y) &= \frac{1}{4 b^2} \int_{\mu_2}^{\infty} \int_{\mu_1}^{\mu_2} (y - x) \cdot e^{\frac{-(x - \mu_1)}{b}} \cdot e^{\frac{-(y - \mu_2)}{b}} \, dx\, dy\\
 &= \frac{1}{4 b^2} \int_{\mu_2}^{\infty} \left( \int_{\mu_1}^{\mu_2} (y - x) \cdot e^{\frac{-(x - \mu_1)}{b}} \, dx \right) \cdot e^{\frac{-(y - \mu_2)}{b}} \, dy\\
 &= \frac{1}{4 b^2} \int_{\mu_2}^{\infty} \left( b y - b \mu_1 - b^2 - e^\frac{\mu_1 - \mu_2}{b} \left(b y - b \mu_2 - b^2 \right) \right) \cdot e^{\frac{-(y - \mu_2)}{b}} \, dy\\
 &= \frac{1}{4 b} \int_{\mu_2}^{\infty} \left( y - \mu_1 - b - e^\frac{\mu_1 - \mu_2}{b} \left(y - \mu_2 - b \right) \right) \cdot e^{\frac{-(y - \mu_2)}{b}} \, dy\\
 &= \frac{1}{4 b} \left( b \mu_2 - b \mu_1 \right) \\
 &= \frac{1}{4} \left( \mu_2 - \mu_1 \right)
\end{align*}

\subsection{$\mu_2 < x, y \leq \mu_1$}

Rewriting Equation~\ref{eq:laplace_lk_symmetric_b} for the part of the space where $\mu_2 < x, y < \mu_1$ is
\begin{align*}
d_{\mu_2 < x, y \leq \mu_1}(X, Y) &= \frac{1}{4 b^2} \int_{-\infty}^{\mu_1} \int_{\mu_2}^{\infty} (x - y) \cdot e^{\frac{-(x - \mu_1)}{b}} \cdot e^{\frac{-(\mu_2 - y)}{b}} \, dx\, dy\\
 &= \frac{1}{4 b^2} \int_{-\infty}^{\mu_1} \left( \int_{\mu_2}^{\infty} (x - y) \cdot e^{\frac{-(x - \mu_1)}{b}} \, dx \right) \cdot e^{\frac{-(\mu_2 - y)}{b}} \, dy\\
 &= \frac{1}{4 b^2} \int_{-\infty}^{\mu_1} e^\frac{\mu_1 - \mu_2}{b} \left( -b y + b \mu_2 + b^2 \right) \cdot e^{\frac{-(\mu_2 - y)}{b}} \, dy\\
 &= \frac{1}{4 b} \int_{-\infty}^{\mu_1} e^\frac{\mu_1 - \mu_2}{b} \left( -y + \mu_2 + b \right) \cdot e^{\frac{-(\mu_2 - y)}{b}} \, dy\\
 &= \frac{1}{4 b} e^\frac{2 \mu_1 - 2 \mu_2}{b} \left( b \mu_2 + 2 b^2 - b \mu_1 \right)\\
 &= \frac{1}{4} e^\frac{2 \mu_1 - 2 \mu_2}{b} \left( \mu_2 + 2 b - \mu_1 \right)\\
 &= \frac{1}{4} e^\frac{-2 (\mu_2 - \mu_1)}{b} \left(2 b + \mu_2 - \mu_1 \right)
\end{align*}

\subsection{$\mu_2 < x, \mu_1 < y \leq \mu_2$}

Rewriting Equation~\ref{eq:laplace_lk_symmetric_b} for the part of the space where $\mu_2 < x, \mu_1 < y < \mu_2$ is
\begin{align*}
d_{\mu_2 < x, \mu_1 < y \leq \mu_2}(X, Y) &= \frac{1}{4 b^2} \int_{\mu_1}^{\mu_2} \int_{\mu_2}^{\infty} (x - y) \cdot e^{\frac{-(x - \mu_1)}{b}} \cdot e^{\frac{-(\mu_2 - y)}{b}} \, dx\, dy\\
 &= \frac{1}{4 b^2} \int_{\mu_1}^{\mu_2} \left( \int_{\mu_2}^{\infty} (x - y) \cdot e^{\frac{-(x - \mu_1)}{b}} \, dx \right) \cdot e^{\frac{-(\mu_2 - y)}{b}} \, dy\\
 &= \frac{1}{4 b^2} \int_{\mu_1}^{\mu_2} \left( e^\frac{\mu_1-\mu_2}{b} \left( -b y + b \mu_2 + b^2 \right) \right) \cdot e^{\frac{-(\mu_2 - y)}{b}} \, dy\\
 &= \frac{1}{4 b} \int_{\mu_1}^{\mu_2} \left( e^\frac{\mu_1-\mu_2}{b} \left( -y + \mu_2 + b \right) \right) \cdot e^{\frac{-(\mu_2 - y)}{b}} \, dy\\
 &= \frac{1}{4 b} \left( 2 b^2 e^\frac{\mu_1 - \mu_2}{b} - e^\frac{2 \mu_1 - 2 \mu_2}{b} \left( b \mu_2 - b \mu_1 + 2 b^2 \right) \right) \\
 &= \frac{1}{4} \left( 2 b e^\frac{\mu_1 - \mu_2}{b} + e^\frac{2 \mu_1 - 2 \mu_2}{b} \left( \mu_1 - \mu_2 - 2 b \right) \right)
\end{align*}

% \subsection{$\mu_2 < x, \mu_2 < y$}

Rewriting Equation~\ref{eq:laplace_lk_symmetric_b} for the part of the space where $\mu_2 < x, \mu_2 < y$ is
\begin{align*}
d_{\mu_2 < x, \mu_2 < y}(X, Y) &= \frac{1}{4 b^2} \int_{\mu_2}^{\infty} \int_{\mu_2}^{\infty} |x - y| \cdot e^{\frac{-(x - \mu_1)}{b}} \cdot e^{\frac{-(y - \mu_2)}{b}} \, dx\, dy\\
 &= \frac{1}{4 b^2} \int_{\mu_2}^{\infty} \left( \int_{\mu_2}^{\infty} |x - y| \cdot e^{\frac{-(x - \mu_1)}{b}} \, dx \right) \cdot e^{\frac{-(y - \mu_2)}{b}} \, dy\\
 &= \frac{1}{4 b^2} \int_{\mu_2}^{\infty} \left( \int_{\mu_2}^{y} (y - x) \cdot e^{\frac{-(x - \mu_1)}{b}} \, dx + \int_{y}^{\infty} (x - y) \cdot e^{\frac{-(x - \mu_1)}{b}} \, dx \right) \cdot e^{\frac{-(y - \mu_2)}{b}} \, dy\\
 &= \frac{1}{4 b^2} \int_{\mu_2}^{\infty} \left( \left( b^2 e^\frac{\mu_1 - y}{b} + e^\frac{\mu_1 - \mu_2}{b} \left(b y - b \mu_2 - b^2 \right) \right) + \left( b^2 e^\frac{\mu_1 - y}{b} \right) \right) \cdot e^{\frac{-(y - \mu_2)}{b}} \, dy\\
 &= \frac{1}{4 b} \int_{\mu_2}^{\infty} \left( 2 b e^\frac{\mu_1 - y}{b} + e^\frac{\mu_1 - \mu_2}{b} \left(y - \mu_2 - b \right) \right) \cdot e^{\frac{-(y - \mu_2)}{b}} \, dy\\
 &= \frac{1}{4 b} b^2 e^\frac{\mu_1 - \mu_2}{b}\\
 &= \frac{1}{4} b e^\frac{-(\mu_2 - \mu_1)}{b}.
\end{align*}

\subsection{Combining the Parts}

We can combine each of the probability weighted distances as
\begin{align*}
d(X, Y) &= d_{x \leq \mu_1, y \leq \mu_1}(X, Y) + d_{x \leq \mu_1, \mu_1 < y \leq \mu_2}(X, Y) + d_{x \leq \mu_1, \mu_2 < y}(X, Y) + d_{\mu_1 < x \leq \mu_2, y \leq \mu_1}(X, Y) \\
&\quad + d_{\mu_1 < x \leq \mu_2, \mu_1 < y \leq \mu_2}(X, Y) + d_{\mu_1 < x \leq \mu_2, \mu_2 < y}(X, Y) + d_{\mu_2 < x, y \leq \mu_1}(X, Y) + d_{\mu_2 < x, \mu_1 < y \leq \mu_2}(X, Y) \\
&\quad + d_{\mu_2 < x, \mu_2 < y}(X, Y)\\
&= \frac{1}{4} b e^\frac{-(\mu_2 - \mu_1)}{b} \\
&\quad+ \frac{1}{4} \left( \mu_2 - \mu_1 \right) \\
&\quad+ \frac{1}{4} \left( \mu_2 - \mu_1 + 2 b \right) \\
&\quad+ \frac{1}{4} \left( 2 b e^\frac{\mu_1-\mu_2}{b} + e^\frac{2 \mu_1 - 2 \mu_2}{b} \left( \mu_1 - \mu_2 - 2 b \right) \right) \\
&\quad+ \frac{1}{4} \left( \mu_2 - \mu_1 - 2 b + e^\frac{\mu_1 - \mu_2}{b} \left( 2 \mu_2 - 2 \mu_1 \right) + e^\frac{2 \mu_1 - 2 \mu_2}{b} \left(\mu_2 - \mu_1 + 2 b \right) \right)\\
&\quad+ \frac{1}{4} \left( \mu_2 - \mu_1 \right) \\
&\quad+ \frac{1}{4} e^\frac{-2 (\mu_2 - \mu_1)}{b} \left(2 b + \mu_2 - \mu_1 \right) \\
&\quad+ \frac{1}{4} \left( 2 b e^\frac{\mu_1 - \mu_2}{b} + e^\frac{2 \mu_1 - 2 \mu_2}{b} \left( \mu_1 - \mu_2 - 2 b \right) \right) \\
&\quad+ \frac{1}{4} b e^\frac{-(\mu_2 - \mu_1)}{b} \\
&= b e^\frac{-(\mu_2 - \mu_1)}{b} \\
&\quad+ \frac{1}{2} \left( \mu_2 - \mu_1 + b \right) \\
&\quad+ \frac{1}{4} \left( 2 b e^\frac{\mu_1-\mu_2}{b} + e^\frac{2 \mu_1 - 2 \mu_2}{b} \left( \mu_1 - \mu_2 - 2 b \right) \right) \\
&\quad+ \frac{1}{4} \left( \mu_2 - \mu_1 - 2 b + e^\frac{\mu_1 - \mu_2}{b} \left( 2 \mu_2 - 2 \mu_1 \right) + e^\frac{2 \mu_1 - 2 \mu_2}{b} \left(\mu_2 - \mu_1 + 2 b \right) \right) \\
&\quad+ \frac{1}{4} \left( \mu_2 - \mu_1 \right) \\
&= b e^\frac{-(\mu_2 - \mu_1)}{b} \\
&\quad+ \left( \mu_2 - \mu_1 \right) \\
&\quad+ \frac{1}{4} \left( 2 b e^\frac{\mu_1-\mu_2}{b} \right) \\
&\quad+ \frac{1}{4} \left( e^\frac{\mu_1 - \mu_2}{b} \left( 2 \mu_2 - 2 \mu_1 \right)  \right) \\
&= b e^\frac{-(\mu_2 - \mu_1)}{b} + \left( \mu_2 - \mu_1 \right) + \frac{1}{2} b e^\frac{\mu_1-\mu_2}{b} + \frac{1}{2} e^\frac{\mu_1 - \mu_2}{b} \left( \mu_2 - \mu_1 \right) \\
d(X, Y) &= \mu_2 - \mu_1 + \frac{1}{2} e^\frac{-(\mu_2 - \mu_1)}{b} \left( 3 b + \mu_2 - \mu_1 \right).
\end{align*}

To remove the assumption that $\mu_1 \leq \mu_2$, we can rewrite this result as
\begin{equation}
\label{eq:lk_laplace_}
d(X, Y) = |\mu_2 - \mu_1| + \frac{1}{2} e^\frac{-|\mu_2 - \mu_1|}{b} \left( 3 b + |\mu_2 - \mu_1| \right). 
\end{equation}

This completes the derivation of LK distance with Laplace Distributions.

% \subsection{Two exponents (existing solution)}

% From Łukaszyk's dissertation, the solution for two exponential distributions is
% \begin{equation}
% d(x, y) = |\mu_1 - \mu_2| + \frac{2 s^2}{|\mu_1 - \mu_2| + 2 s}.
% \end{equation}

% \subsection{Multiple dimensions with $L_0$}

% in 2 dimensions with $L_0$:

% Given points $s$ and $t$ comprised of two dimensions each, Let $x = |\mu_2 - \mu_1|$ on the first dimension and $y = |\mu_2 - \mu_1|$ on the second dimension, we have
% \begin{equation}
% d(s, t) = \left( \left(x + \frac{1}{2} e^\frac{-x}{b} \left( 3 b + x \right) \right) \left(y + \frac{1}{2} e^\frac{-y}{b} \left( 3 b + y \right) \right) \right)^\frac{1}{2}
% \end{equation}

% The derivative of the Laplace LK metric on each dimension is
% \begin{equation}
% \frac{d}{dx} \left(x + \frac{1}{2} e^\frac{-x}{b} \left( 3 b + x \right) \right) = 1 + \frac{1}{2} e^\frac{-x}{b} - \frac{1}{2 b} \left( x + 3 b \right) e^\frac{-x}{b}.
% \end{equation}

% To find the extrema of the LK metric, we set the derivative to zero and rearrange to find
% \begin{equation}
% x \frac{e^\frac{-x}{b}}{1 - e^\frac{-x}{b}} = 2 b.
% \end{equation}

\newpage

% \subsection{}

\section{Information about the Benchmarked Algorithms}
\begin{itemize}
    \item Gradient Boosted Trees
    
    K-fold cross-validation was carried out with $K=6$. Grid search was carried out on the number of estimators ($N_{\text{est}}$) such that $N_{\text{est}} \in \{ \lceil e^i \rceil \}_{i=1}^8$.  
    
    \item Tradional KNN

    K-fold cross-validation was carried out with $K=6$. Grid search was carried out on the number of the number of neighbors ($k_n$) and the value of $p$ in $l_p$ norms such that $k_n \in [1, 2, 3, 5, 8, 13, 21, 34, 55, 89, 144]$ and $p \in [1, 2]$. Note that the search space of neighbors are picked according to the Fibonacci sequence since it grows at a slower rate than some other sequences (e.g. exponential), which is advantageous when exploring hyperparameter values. It provides a versatile set of values that can adapt to different datasets and problem characteristics leading to a more diverse exploration of the search space, helping to identify a wider range of potential optimal $k_n$ values.
    
    \item Regularized Least Squares
    
     Elastic Net was used in the case of regression. K-fold cross-validation was carried out with $K=6$. Grid search was carried out on the scaling ration of the $l_1$ and $l_2$ penalties ranging from values $[0.1, 0.5, 0.7, 0.9, 0.95, 0.99, 1.0]$
     
    \item Logistic Regression

    K-fold cross-validation was carried out with $K=6$. Grid search was carried out on the inverse of regularization strength $C$ in the logarithmic scale such that $$C \in [0.0001, 0.0008, 0.006, 0.0464, 0.3594, 2.7826, 21.5443, 166.8101, 1291.5497, 10000.0]$$ The optimization problem was solved using stochastic average gradient with $l_2$ penalty.
    
    \item Neural Network
    
     For both classification and regression datasets, Adam Optimizer was used with batch size 128 and learning rate of $0.001$. An internal validation set of $10\%$ was used from the training data for an early stopping criteria, with maximum epochs set to $1000$ using swish activation for each hidden layer. A dropout rate of $10\%$ with Layer Norm was used after each hidden layer. The details of the architecture can be found in the table below.

    \begin{table*}[h]
        \centering
    \scriptsize{
\begin{tabular}{lcl}
    \hline
       Layer \# & Parameter Type   & Parameter Size   \\
    \hline
     layers 1       & 0.weight         & (512, Input Size)        \\
     layers 1     & 0.bias           & (512,)           \\
     dropout 1     & 1.weight         & (512,)           \\
     dropout 1     & 1.bias           & (512,)           \\
     layers 2      & 4.weight         & (512, 512)       \\
     layers 2      & 4.bias           & (512,)           \\
     dropout 2      & 5.weight         & (512,)           \\
     dropout 2      & 5.bias           & (512,)           \\
     layers 3      & 8.weight         & (512, 512)       \\
     layers 3      & 8.bias           & (512,)           \\
     dropout 3      & 9.weight         & (512,)           \\
     dropout 3      & 9.bias           & (512,)           \\
     Output Layer & weight           & (Output Size, 512)         \\
     Output Layer & bias             & (Output Size,)             \\
    \hline
    \end{tabular}
        \label{tab:nn_arch}
        }
    \end{table*}
    
    \item Random Forests

    K-fold cross-validation was carried out with $K=6$. Grid search was carried out on the number of estimators ($N_{\text{est}}$) such that $N_{\text{est}} \in \{ \lceil e^i \rceil \}_{i=1}^8$.  
    
    \item Light-GBM
    
    K-fold cross-validation was carried out with $K=6$. The number of estimators used was 100 with number of leaves = 31

\end{itemize}

\begin{table}[!ht]
    \centering
\scriptsize{
\section{Detailed Information about the PMLB Classificaton Datasets}
\vskip 0.15in
% \begin{sc}
\begin{tabular}{lcccc}
\toprule
{} &                      Dataset Name &       Rows &    Columns &       Rows $\times$ Columns\\
\midrule
1   &        GAMETES\_Epistasis\_2\_Way\_20atts\_0.1H\_EDM\_1\_1 &   1600 &   21 &   33600 \\
2   &        GAMETES\_Epistasis\_2\_Way\_20atts\_0.4H\_EDM\_1\_1 &   1600 &   21 &   33600 \\
3   &        GAMETES\_Epistasis\_3\_Way\_20atts\_0.2H\_EDM\_1\_1 &   1600 &   21 &   33600 \\
4   &  GAMETES\_Heterogeneity\_20atts\_1600\_Het\_0.4\_0.2\_... &   1600 &   21 &   33600 \\
5   &  GAMETES\_Heterogeneity\_20atts\_1600\_Het\_0.4\_0.2\_... &   1600 &   21 &   33600 \\
6   &                             Hill\_Valley\_with\_noise &   1212 &  101 &  122412 \\
7   &                          Hill\_Valley\_without\_noise &   1212 &  101 &  122412 \\
\bottomrule
\end{tabular}
    \label{fig:class_shapes_1}
    }
\end{table}
% 400000

\begin{table*}[!ht]
    \centering
\scriptsize{
\begin{tabular}{lcccc}
\toprule
{} &                      Dataset Name &       Rows &    Columns &       Rows $\times$ Columns\\
\midrule
74  &                                            heart\_h &    294 &   14 &    4116 \\
75  &                                      heart\_statlog &    270 &   14 &    3780 \\
76  &                                          hepatitis &    155 &   20 &    3100 \\
77  &                                        horse\_colic &    368 &   23 &    8464 \\
78  &                                     house\_votes\_84 &    435 &   17 &    7395 \\
79  &                                          hungarian &    294 &   14 &    4116 \\
80  &                                        hypothyroid &   3163 &   26 &   82238 \\
\bottomrule
\end{tabular}
    \label{fig:class_shapes_2}
    }
\end{table*}

% Regression data info

\begin{table*}[!ht]
    \centering
\scriptsize{
\section{Detailed Information about the PMLB Regression Datasets}
\vskip 0.15in
\begin{tabular}{lcccc}
\toprule
{} &                      Dataset Name &       Rows &    Columns &       Rows $\times$ Columns\\
\midrule
1   &                          1027\_ESL &     488 &    5 &    2440 \\
2   &                          1028\_SWD &    1000 &   11 &   11000 \\
3   &                          1029\_LEV &    1000 &    5 &    5000 \\
4   &                          1030\_ERA &    1000 &    5 &    5000 \\
5   &                      1089\_USCrime &      47 &   14 &     658 \\
6   &              1096\_FacultySalaries &      50 &    5 &     250 \\
7   &               1199\_BNG\_echoMonths &   17496 &   10 &  174960 \\
\bottomrule
\end{tabular}
    \label{fig:reg_shapes_1}
    }
\end{table*}

\begin{table*}[!ht]
    \centering
\scriptsize{
\begin{tabular}{lcccc}
\toprule
{} &                      Dataset Name &       Rows &    Columns &       Rows $\times$ Columns\\
\midrule
55  &                 612\_fri\_c1\_1000\_5 &    1000 &    6 &    6000 \\
56  &                  613\_fri\_c3\_250\_5 &     250 &    6 &    1500 \\
57  &                 615\_fri\_c4\_250\_10 &     250 &   11 &    2750 \\
58  &                 616\_fri\_c4\_500\_50 &     500 &   51 &   25500 \\
59  &                  617\_fri\_c3\_500\_5 &     500 &    6 &    3000 \\
60  &                618\_fri\_c3\_1000\_50 &    1000 &   51 &   51000 \\
61  &                620\_fri\_c1\_1000\_25 &    1000 &   26 &   26000 \\
\bottomrule
\end{tabular}
    \label{fig:reg_shapes_2}
    }
\end{table*}

\begin{table*}[!ht]
    \centering
\scriptsize{
\begin{tabular}{lcccc}
\toprule
{} &                      Dataset Name &       Rows &    Columns &       Rows $\times$ Columns\\
\midrule
109 &                   feynman\_II\_15\_5 &  100000 &    4 &  400000 \\
110 &                  feynman\_II\_24\_17 &  100000 &    4 &  400000 \\
111 &                  feynman\_II\_27\_16 &  100000 &    4 &  400000 \\
112 &                  feynman\_II\_27\_18 &  100000 &    3 &  300000 \\
113 &                   feynman\_II\_34\_2 &  100000 &    4 &  400000 \\
114 &                 feynman\_II\_34\_29a &  100000 &    4 &  400000 \\
115 &                  feynman\_II\_34\_2a &  100000 &    4 &  400000 \\
\bottomrule
\end{tabular}
    \label{fig:reg_shapes_3}
    }
\end{table*}

\begin{table*}[!ht]
\section{Detailed Results: Classification}
\vskip 0.15in
    \centering
\scriptsize{
\begin{tabular}{ccccccc}
\toprule
{} &                                       Dataset Name & Algorithm & Mean Accuracy [0-1] $\uparrow$ & Mean Precision $\uparrow$ & Mean Recall $\uparrow$ &  Mean MCC $\uparrow$  \\
\midrule
1   &        GAMETES\_Epistasis\_2\_Way\_20atts\_0.1H\_EDM\_1\_1 &      Ours &      0.653854 &       0.669697 &    0.655529 &   0.32489 \\
2   &        GAMETES\_Epistasis\_2\_Way\_20atts\_0.4H\_EDM\_1\_1 &      Ours &      0.759479 &       0.777411 &    0.761283 &  0.538387 \\
3   &        GAMETES\_Epistasis\_3\_Way\_20atts\_0.2H\_EDM\_1\_1 &      Ours &      0.659583 &       0.662087 &    0.660329 &  0.322405 \\
4   &  GAMETES\_Heterogeneity\_20atts\_1600\_Het\_0.4\_0.2\_... &      Ours &      0.691042 &       0.692929 &    0.691785 &  0.384708 \\
5   &  GAMETES\_Heterogeneity\_20atts\_1600\_Het\_0.4\_0.2\_... &      Ours &      0.671354 &       0.674906 &    0.672593 &  0.347478 \\
6   &                             Hill\_Valley\_with\_noise &      Ours &      0.555693 &        0.56096 &    0.555904 &  0.116711 \\
7   &                          Hill\_Valley\_without\_noise &      Ours &      0.634979 &        0.63918 &    0.635376 &  0.274511 \\
\bottomrule
\end{tabular}
    \label{fig:class_full_1}
    }
\end{table*}

\begin{table*}[!ht]
    \centering
\scriptsize{
\begin{tabular}{ccccccc}
\toprule
{} &                                       Dataset Name & Algorithm & Mean Accuracy [0-1] $\uparrow$ & Mean Precision $\uparrow$ & Mean Recall $\uparrow$ &  Mean MCC $\uparrow$  \\
\midrule

74  &                                            heart\_h &      Ours &       0.79661 &       0.791894 &    0.765024 &  0.555181 \\
75  &                                      heart\_statlog &      Ours &      0.828395 &       0.832279 &    0.818623 &  0.650413 \\
76  &                                          hepatitis &      Ours &      0.815054 &       0.716643 &    0.635302 &  0.335239 \\
77  &                                        horse\_colic &      Ours &       0.82027 &       0.814153 &    0.801003 &  0.614658 \\
78  &                                     house\_votes\_84 &      Ours &       0.95249 &       0.948284 &    0.952441 &  0.900652 \\
79  &                                          hungarian &      Ours &      0.812429 &       0.803013 &    0.781021 &  0.582977 \\
80  &                                        hypothyroid &      Ours &      0.972617 &       0.889952 &    0.795092 &  0.675621 \\
\bottomrule
\end{tabular}
    \label{fig:class_full_ours2}
    }
\end{table*}

% GRADIENT BOOST--------------------------------------------------------------------------------------------------------------------

\begin{table*}[!ht]
    \centering
\scriptsize{
\begin{tabular}{ccccccc}
\toprule
{} &                                       Dataset Name & Algorithm & Mean Accuracy [0-1] $\uparrow$ & Mean Precision $\uparrow$ & Mean Recall $\uparrow$ &  Mean MCC $\uparrow$  \\
\midrule
1   &        GAMETES\_Epistasis\_2\_Way\_20atts\_0.1H\_EDM\_1\_1 &        GB &      0.623229 &       0.624565 &    0.623842 &  0.248402 \\
2   &        GAMETES\_Epistasis\_2\_Way\_20atts\_0.4H\_EDM\_1\_1 &        GB &      0.726458 &       0.728732 &    0.726848 &  0.455558 \\
3   &        GAMETES\_Epistasis\_3\_Way\_20atts\_0.2H\_EDM\_1\_1 &        GB &      0.538438 &       0.538718 &    0.538705 &  0.077424 \\
4   &  GAMETES\_Heterogeneity\_20atts\_1600\_Het\_0.4\_0.2\_... &        GB &      0.681458 &       0.682047 &      0.6818 &  0.363846 \\
5   &  GAMETES\_Heterogeneity\_20atts\_1600\_Het\_0.4\_0.2\_... &        GB &      0.699792 &       0.700784 &    0.700226 &  0.401008 \\
6   &                             Hill\_Valley\_with\_noise &        GB &      0.554321 &       0.555855 &    0.555435 &  0.111285 \\
7   &                          Hill\_Valley\_without\_noise &        GB &      0.592044 &       0.592688 &    0.592331 &  0.185017 \\
\bottomrule
\end{tabular}
    \label{fig:class_full_gb1}
    }
\end{table*}

\begin{table*}[!ht]
    \centering
\scriptsize{
\begin{tabular}{ccccccc}
\toprule
{} &                                       Dataset Name & Algorithm & Mean Accuracy [0-1] $\uparrow$ & Mean Precision $\uparrow$ & Mean Recall $\uparrow$ &  Mean MCC $\uparrow$  \\
\midrule
74  &                                            heart\_h &        GB &      0.810734 &       0.802262 &    0.788048 &  0.589686 \\
75  &                                      heart\_statlog &        GB &      0.809259 &       0.810886 &    0.801569 &  0.611764 \\
76  &                                          hepatitis &        GB &      0.809677 &       0.679517 &    0.615463 &  0.297586 \\
77  &                                        horse\_colic &        GB &      0.825676 &       0.824893 &    0.803777 &  0.627761 \\
78  &                                     house\_votes\_84 &        GB &      0.956322 &       0.952215 &    0.956129 &  0.908305 \\
79  &                                          hungarian &        GB &      0.822599 &       0.812434 &    0.796415 &  0.608083 \\
80  &                                        hypothyroid &        GB &      0.960295 &       0.821943 &    0.690758 &  0.492934 \\
\bottomrule
\end{tabular}
    \label{fig:class_full_gb2}
    }
\end{table*}

% KNN--------------------------------------------------------------------------------------------------------------------

\begin{table*}[!ht]
    \centering
\scriptsize{
\begin{tabular}{ccccccc}
\toprule
{} &                                       Dataset Name & Algorithm & Mean Accuracy [0-1] $\uparrow$ & Mean Precision $\uparrow$ & Mean Recall $\uparrow$ &  Mean MCC $\uparrow$  \\
\midrule
1   &        GAMETES\_Epistasis\_2\_Way\_20atts\_0.1H\_EDM\_1\_1 &       KNN &      0.575208 &       0.580815 &    0.576832 &  0.157566 \\
2   &        GAMETES\_Epistasis\_2\_Way\_20atts\_0.4H\_EDM\_1\_1 &       KNN &      0.700521 &       0.707414 &    0.702078 &  0.409406 \\
3   &        GAMETES\_Epistasis\_3\_Way\_20atts\_0.2H\_EDM\_1\_1 &       KNN &       0.54875 &       0.550881 &    0.549925 &  0.100793 \\
4   &  GAMETES\_Heterogeneity\_20atts\_1600\_Het\_0.4\_0.2\_... &       KNN &      0.635208 &       0.637862 &    0.636591 &  0.274444 \\
5   &  GAMETES\_Heterogeneity\_20atts\_1600\_Het\_0.4\_0.2\_... &       KNN &      0.641563 &       0.644375 &    0.642843 &  0.287202 \\
6   &                             Hill\_Valley\_with\_noise &       KNN &       0.58011 &        0.58036 &    0.580044 &  0.160401 \\
7   &                          Hill\_Valley\_without\_noise &       KNN &      0.625652 &       0.626107 &    0.625637 &  0.251741 \\
\bottomrule
\end{tabular}
    \label{fig:class_full_knn1}
    }
\end{table*}

\begin{table*}[!ht]
    \centering
\scriptsize{
\begin{tabular}{ccccccc}
\toprule
{} &                                       Dataset Name & Algorithm & Mean Accuracy [0-1] $\uparrow$ & Mean Precision $\uparrow$ & Mean Recall $\uparrow$ &  Mean MCC $\uparrow$  \\
\midrule
74  &                                            heart\_h &       KNN &       0.79096 &       0.785242 &    0.758757 &  0.542648 \\
75  &                                      heart\_statlog &       KNN &      0.701235 &       0.699152 &    0.692409 &  0.391261 \\
76  &                                          hepatitis &       KNN &      0.823656 &       0.777807 &    0.633933 &  0.371221 \\
77  &                                        horse\_colic &       KNN &           0.8 &       0.796235 &    0.773937 &  0.569391 \\
78  &                                     house\_votes\_84 &       KNN &      0.918774 &       0.912232 &    0.919218 &  0.831325 \\
79  &                                          hungarian &       KNN &      0.811299 &       0.802042 &    0.776375 &  0.577313 \\
80  &                                        hypothyroid &       KNN &      0.957609 &       0.890294 &    0.604244 &   0.38972 \\
\bottomrule
\end{tabular}
    \label{fig:class_full_knn2}
    }
\end{table*}

% LR--------------------------------------------------------------------------------------------------------------------

\begin{table*}[!ht]
    \centering
\scriptsize{
\begin{tabular}{ccccccc}
\toprule
{} &                                       Dataset Name & Algorithm & Mean Accuracy [0-1] $\uparrow$ & Mean Precision $\uparrow$ & Mean Recall $\uparrow$ &  Mean MCC $\uparrow$  \\
\midrule
1   &        GAMETES\_Epistasis\_2\_Way\_20atts\_0.1H\_EDM\_1\_1 &    Linear &      0.480938 &       0.458951 &     0.48729 & -0.028143 \\
2   &        GAMETES\_Epistasis\_2\_Way\_20atts\_0.4H\_EDM\_1\_1 &    Linear &      0.473333 &       0.496988 &    0.479564 & -0.039612 \\
3   &        GAMETES\_Epistasis\_3\_Way\_20atts\_0.2H\_EDM\_1\_1 &    Linear &      0.505729 &       0.507804 &    0.507221 &  0.014985 \\
4   &  GAMETES\_Heterogeneity\_20atts\_1600\_Het\_0.4\_0.2\_... &    Linear &      0.478437 &        0.48032 &    0.485211 & -0.030165 \\
5   &  GAMETES\_Heterogeneity\_20atts\_1600\_Het\_0.4\_0.2\_... &    Linear &      0.484792 &       0.479447 &    0.490373 &  -0.01969 \\
6   &                             Hill\_Valley\_with\_noise &    Linear &      0.750754 &       0.822602 &    0.752891 &  0.570721 \\
7   &                          Hill\_Valley\_without\_noise &    Linear &      0.707133 &       0.810475 &    0.709951 &  0.510171 \\
\bottomrule
\end{tabular}
    \label{fig:class_full_lr1}
    }
\end{table*}

\begin{table*}[!ht]
    \centering
\scriptsize{
\begin{tabular}{ccccccc}
\toprule
{} &                                       Dataset Name & Algorithm & Mean Accuracy [0-1] $\uparrow$ & Mean Precision $\uparrow$ & Mean Recall $\uparrow$ &  Mean MCC $\uparrow$  \\
\midrule
74  &                                            heart\_h &    Linear &      0.819774 &        0.81881 &    0.789051 &  0.606259 \\
75  &                                      heart\_statlog &    Linear &      0.842593 &        0.84248 &    0.835538 &  0.677814 \\
76  &                                          hepatitis &    Linear &      0.825806 &       0.718776 &    0.661382 &  0.373473 \\
77  &                                        horse\_colic &    Linear &      0.813964 &         0.8109 &    0.789035 &  0.599212 \\
78  &                                     house\_votes\_84 &    Linear &      0.952107 &       0.948088 &    0.951188 &  0.899222 \\
79  &                                          hungarian &    Linear &      0.837853 &       0.835924 &    0.803843 &  0.638102 \\
80  &                                        hypothyroid &    Linear &      0.972828 &       0.879405 &    0.809378 &  0.683784 \\
\bottomrule
\end{tabular}
    \label{fig:class_full_lr2}
    }
\end{table*}

% Neural Net--------------------------------------------------------------------------------------------------------------------

\begin{table*}[!ht]
    \centering
\scriptsize{
\begin{tabular}{ccccccc}
\toprule
{} &                                       Dataset Name & Algorithm & Mean Accuracy [0-1] $\uparrow$ & Mean Precision $\uparrow$ & Mean Recall $\uparrow$ &  Mean MCC $\uparrow$  \\
\midrule
1   &        GAMETES\_Epistasis\_2\_Way\_20atts\_0.1H\_EDM\_1\_1 &  NeuralNet &       0.585729 &        0.585261 &     0.585838 &  0.171095 \\
2   &        GAMETES\_Epistasis\_2\_Way\_20atts\_0.4H\_EDM\_1\_1 &  NeuralNet &       0.708542 &        0.708453 &     0.709424 &  0.417870 \\
3   &        GAMETES\_Epistasis\_3\_Way\_20atts\_0.2H\_EDM\_1\_1 &  NeuralNet &       0.517500 &        0.517558 &     0.517636 &  0.035193 \\
4   &  GAMETES\_Heterogeneity\_20atts\_1600\_Het\_0.4\_0.2\_... &  NeuralNet &       0.644687 &        0.645142 &     0.646459 &  0.291590 \\
5   &  GAMETES\_Heterogeneity\_20atts\_1600\_Het\_0.4\_0.2\_... &  NeuralNet &       0.650833 &        0.651469 &     0.652644 &  0.304104 \\
6   &                             Hill\_Valley\_with\_noise &  NeuralNet &       0.545405 &        0.550122 &     0.610026 &  0.146337 \\
7   &                          Hill\_Valley\_without\_noise &  NeuralNet &       0.554458 &        0.560038 &     0.622925 &  0.168091 \\
\bottomrule
\end{tabular}
    \label{fig:class_full_nn1}
    }
\end{table*}

\begin{table*}[!ht]
    \centering
\scriptsize{
\begin{tabular}{ccccccc}
\toprule
{} &                                       Dataset Name & Algorithm & Mean Accuracy [0-1] $\uparrow$ & Mean Precision $\uparrow$ & Mean Recall $\uparrow$ &  Mean MCC $\uparrow$  \\
\midrule
74  &                                            heart\_h &  NeuralNet &       0.774576 &        0.745714 &     0.767058 &  0.511602 \\
75  &                                      heart\_statlog &  NeuralNet &       0.801235 &        0.799560 &     0.798810 &  0.598298 \\
76  &                                          hepatitis &  NeuralNet &       0.816129 &        0.691785 &     0.717233 &  0.404004 \\
77  &                                        horse\_colic &  NeuralNet &       0.791441 &        0.767537 &     0.787856 &  0.554577 \\
78  &                                     house\_votes\_84 &  NeuralNet &       0.949425 &        0.948397 &     0.945097 &  0.893421 \\
79  &                                          hungarian &  NeuralNet &       0.793785 &        0.770260 &     0.781005 &  0.550523 \\
80  &                                        hypothyroid &  NeuralNet &       0.950395 &        0.500000 &     0.475197 &  0.000000 \\
\bottomrule
\end{tabular}
    \label{fig:class_full_nn2}
    }
\end{table*}

% Random Forest--------------------------------------------------------------------------------------------------------------------

\begin{table*}[!ht]
    \centering
\scriptsize{
\begin{tabular}{ccccccc}
\toprule
{} &                                       Dataset Name & Algorithm & Mean Accuracy [0-1] $\uparrow$ & Mean Precision $\uparrow$ & Mean Recall $\uparrow$ &  Mean MCC $\uparrow$  \\
\midrule
1   &        GAMETES\_Epistasis\_2\_Way\_20atts\_0.1H\_EDM\_1\_1 &        RF &      0.592813 &       0.593805 &    0.593487 &   0.18729 \\
2   &        GAMETES\_Epistasis\_2\_Way\_20atts\_0.4H\_EDM\_1\_1 &        RF &      0.716042 &       0.717502 &    0.716862 &  0.434362 \\
3   &        GAMETES\_Epistasis\_3\_Way\_20atts\_0.2H\_EDM\_1\_1 &        RF &      0.569167 &       0.570024 &    0.569777 &    0.1398 \\
4   &  GAMETES\_Heterogeneity\_20atts\_1600\_Het\_0.4\_0.2\_... &        RF &          0.65 &       0.650981 &    0.650655 &  0.301635 \\
5   &  GAMETES\_Heterogeneity\_20atts\_1600\_Het\_0.4\_0.2\_... &        RF &      0.676042 &       0.676914 &    0.676607 &   0.35352 \\
6   &                             Hill\_Valley\_with\_noise &        RF &      0.566667 &       0.567162 &    0.566994 &  0.134155 \\
7   &                          Hill\_Valley\_without\_noise &        RF &      0.591358 &       0.592646 &    0.592079 &   0.18472 \\
\bottomrule
\end{tabular}
    \label{fig:class_full_rf1}
    }
\end{table*}

\begin{table*}[!ht]
    \centering
\scriptsize{
\begin{tabular}{ccccccc}
\toprule
{} &                                       Dataset Name & Algorithm & Mean Accuracy [0-1] $\uparrow$ & Mean Precision $\uparrow$ & Mean Recall $\uparrow$ &  Mean MCC $\uparrow$  \\
\midrule
74  &                                            heart\_h &        RF &      0.812429 &       0.803651 &    0.791494 &  0.594451 \\
75  &                                      heart\_statlog &        RF &      0.824074 &       0.822665 &    0.816857 &  0.639373 \\
76  &                                          hepatitis &        RF &           0.8 &       0.729485 &    0.625241 &  0.325763 \\
77  &                                        horse\_colic &        RF &      0.843243 &       0.847947 &    0.817756 &  0.664626 \\
78  &                                     house\_votes\_84 &        RF &      0.954023 &       0.951141 &    0.952488 &  0.903554 \\
79  &                                          hungarian &        RF &      0.833333 &       0.823948 &    0.807803 &  0.631089 \\
80  &                                        hypothyroid &        RF &      0.961664 &       0.871592 &    0.661778 &  0.486384 \\
\bottomrule
\end{tabular}
    \label{fig:class_full_rf2}
    }
\end{table*}

% LGBM--------------------------------------------------------------------------------------------------------------------

\begin{table*}[!ht]
    \centering
\scriptsize{
\begin{tabular}{ccccccc}
\toprule
{} &                                       Dataset Name & Algorithm & Mean Accuracy [0-1] $\uparrow$ & Mean Precision $\uparrow$ & Mean Recall $\uparrow$ &  Mean MCC $\uparrow$  \\
\midrule
1   &        GAMETES\_Epistasis\_2\_Way\_20atts\_0.1H\_EDM\_1\_1 &      LGBM &      0.600625 &       0.601217 &    0.600913 &   0.20213 \\
2   &        GAMETES\_Epistasis\_2\_Way\_20atts\_0.4H\_EDM\_1\_1 &      LGBM &      0.725521 &       0.727364 &     0.72608 &  0.453438 \\
3   &        GAMETES\_Epistasis\_3\_Way\_20atts\_0.2H\_EDM\_1\_1 &      LGBM &      0.546354 &        0.54645 &    0.546371 &  0.092821 \\
4   &  GAMETES\_Heterogeneity\_20atts\_1600\_Het\_0.4\_0.2\_... &      LGBM &         0.665 &       0.665684 &    0.665486 &   0.33117 \\
5   &  GAMETES\_Heterogeneity\_20atts\_1600\_Het\_0.4\_0.2\_... &      LGBM &      0.682604 &       0.683213 &    0.682965 &  0.366177 \\
6   &                             Hill\_Valley\_with\_noise &      LGBM &      0.580247 &       0.580742 &    0.580552 &  0.161293 \\
7   &                          Hill\_Valley\_without\_noise &      LGBM &      0.594513 &       0.596329 &     0.59553 &  0.191853 \\
\bottomrule
\end{tabular}
    \label{fig:class_full_lbgm2}
    }
\end{table*}

\begin{table*}[!ht]
    \centering
\scriptsize{
\begin{tabular}{ccccccc}
\toprule
{} &                                       Dataset Name & Algorithm & Mean Accuracy [0-1] $\uparrow$ & Mean Precision $\uparrow$ & Mean Recall $\uparrow$ &  Mean MCC $\uparrow$  \\
\midrule
74  &                                            heart\_h &      LGBM &      0.784746 &       0.770482 &    0.769739 &  0.539905 \\
75  &                                      heart\_statlog &      LGBM &      0.812346 &       0.811347 &    0.807426 &  0.618626 \\
76  &                                          hepatitis &      LGBM &      0.811828 &       0.718382 &    0.685423 &  0.393035 \\
77  &                                        horse\_colic &      LGBM &       0.82027 &       0.816101 &    0.801575 &  0.617131 \\
78  &                                     house\_votes\_84 &      LGBM &      0.955172 &       0.951344 &    0.953965 &  0.905267 \\
79  &                                          hungarian &      LGBM &       0.79548 &       0.779852 &    0.771316 &  0.550672 \\
80  &                                        hypothyroid &      LGBM &      0.982096 &       0.919059 &    0.887098 &  0.803823 \\
\bottomrule
\end{tabular}
    \label{fig:class_full_lbgm_2}
    }
\end{table*}

%  ================================================================ REGRESSION ===========================================================

% --------------Ours---------------------------------------------------------------------------------------------------
\begin{table*}[!ht]
\section{Detailed Results: Regression}
\vskip 0.15in
    \centering
\scriptsize{
\begin{tabular}{ccccccc}
\toprule
{} &         Dataset Name & Algorithm &   $R^2$ Mean  $\uparrow$  &  MAE   $\downarrow$  &    MSE  $\downarrow$  &   Spearman coeff. $\uparrow$ \\
\midrule
1   &                          1027\_ESL &      Ours &  0.814821 &    0.32415 &    0.364286 &  0.915777 \\
2   &                          1028\_SWD &      Ours &  0.242729 &   0.453667 &    0.488667 &  0.547945 \\
3   &                          1029\_LEV &      Ours &  0.441602 &     0.4215 &    0.503833 &  0.693376 \\
4   &                          1030\_ERA &      Ours &  0.326374 &      1.253 &       2.581 &  0.533985 \\
5   &                      1089\_USCrime &      Ours &  0.761386 &  13.656667 &  343.763333 &   0.85516 \\
6   &              1096\_FacultySalaries &      Ours &  0.554916 &   1.917929 &    6.254915 &  0.881548 \\
7   &               1199\_BNG\_echoMonths &      Ours &  0.457349 &    8.99762 &  135.514966 &  0.642297 \\
\bottomrule
\end{tabular}
    \label{fig:tab_reg_1}
    }
\end{table*}

\begin{table*}[!ht]
    \centering
\scriptsize{
\begin{tabular}{ccccccc}
\toprule
{} &         Dataset Name & Algorithm &   $R^2$ Mean  $\uparrow$  &  MAE   $\downarrow$  &    MSE  $\downarrow$  &   Spearman coeff. $\uparrow$ \\
\midrule
55  &                 612\_fri\_c1\_1000\_5 &      Ours &  0.929314 &   0.204331 &    0.067296 &  0.961312 \\
56  &                  613\_fri\_c3\_250\_5 &      Ours &  0.879142 &   0.269598 &    0.123383 &  0.919196 \\
57  &                 615\_fri\_c4\_250\_10 &      Ours &  0.821346 &   0.326316 &     0.17978 &  0.879216 \\
58  &                 616\_fri\_c4\_500\_50 &      Ours &  0.860252 &   0.288322 &    0.138104 &  0.915084 \\
59  &                  617\_fri\_c3\_500\_5 &      Ours &  0.902494 &   0.233656 &    0.095288 &  0.944492 \\
60  &                618\_fri\_c3\_1000\_50 &      Ours &   0.89088 &   0.254606 &    0.105341 &  0.924638 \\
61  &                620\_fri\_c1\_1000\_25 &      Ours &  0.896452 &    0.25679 &    0.103944 &  0.943626 \\
\bottomrule
\end{tabular}
    \label{fig:tab_reg_2}
    }
\end{table*}

\begin{table*}[!ht]
    \centering
\scriptsize{
\begin{tabular}{ccccccc}
\toprule
{} &         Dataset Name & Algorithm &   $R^2$ Mean  $\uparrow$  &  MAE   $\downarrow$  &    MSE  $\downarrow$  &   Spearman coeff. $\uparrow$ \\
\midrule
109 &                   feynman\_II\_15\_5 &      Ours &  0.999482 &    0.07829 &    0.014174 &  0.999788 \\
110 &                  feynman\_II\_24\_17 &      Ours &  0.999776 &     0.0083 &    0.000166 &  0.999914 \\
111 &                  feynman\_II\_27\_16 &      Ours &  0.999722 &   0.879617 &    2.226821 &  0.999874 \\
112 &                  feynman\_II\_27\_18 &      Ours &  0.999991 &    0.05168 &    0.006177 &  0.999996 \\
113 &                   feynman\_II\_34\_2 &      Ours &  0.999737 &   0.104771 &    0.024674 &  0.999839 \\
114 &                 feynman\_II\_34\_29a &      Ours &  0.999321 &   0.002835 &    0.000036 &  0.999847 \\
115 &                  feynman\_II\_34\_2a &      Ours &  0.999342 &   0.005635 &    0.000138 &  0.999846 \\
\bottomrule
\end{tabular}
    \label{fig:tab_reg_3}
    }
\end{table*}

% GB-------------------------------------------------------------------------------------------------------------------

\begin{table*}[!ht]
    \centering
\scriptsize{
\begin{tabular}{ccccccc}
\toprule
{} &         Dataset Name & Algorithm &   $R^2$ Mean  $\uparrow$  &  MAE   $\downarrow$  &    MSE  $\downarrow$  &   Spearman coeff. $\uparrow$ \\
\midrule
1   &                          1027\_ESL &        GB &  0.843825 &   0.411648 &    0.306425 &  0.933898 \\
2   &                          1028\_SWD &        GB &   0.40216 &   0.500758 &    0.386582 &  0.629681 \\
3   &                          1029\_LEV &        GB &  0.544847 &   0.487292 &     0.41125 &  0.741157 \\
4   &                          1030\_ERA &        GB &  0.354067 &   1.265373 &    2.476943 &  0.561748 \\
5   &                      1089\_USCrime &        GB &  0.735817 &  15.812908 &  388.658039 &  0.829873 \\
6   &              1096\_FacultySalaries &        GB &  0.655666 &   1.692329 &    5.266757 &  0.867306 \\
7   &               1199\_BNG\_echoMonths &        GB &  0.459803 &   9.042931 &  134.895122 &  0.647074 \\
\bottomrule
\end{tabular}
    \label{fig:tab_reg_gb1}
    }
\end{table*}

\begin{table*}[!ht]
    \centering
\scriptsize{
\begin{tabular}{ccccccc}
\toprule
{} &         Dataset Name & Algorithm &   $R^2$ Mean  $\uparrow$  &  MAE   $\downarrow$  &    MSE  $\downarrow$  &   Spearman coeff. $\uparrow$ \\
\midrule
55  &                 612\_fri\_c1\_1000\_5 &        GB &  0.950666 &   0.167881 &      0.0469 &  0.971885 \\
56  &                  613\_fri\_c3\_250\_5 &        GB &  0.888793 &   0.255482 &    0.112866 &  0.930164 \\
57  &                 615\_fri\_c4\_250\_10 &        GB &  0.825742 &   0.309544 &    0.175729 &  0.899653 \\
58  &                 616\_fri\_c4\_500\_50 &        GB &  0.870546 &   0.266212 &    0.126715 &  0.918094 \\
59  &                  617\_fri\_c3\_500\_5 &        GB &   0.91477 &   0.212436 &    0.083035 &  0.952741 \\
60  &                618\_fri\_c3\_1000\_50 &        GB &  0.917416 &   0.217763 &    0.079267 &  0.941898 \\
61  &                620\_fri\_c1\_1000\_25 &        GB &  0.930267 &    0.20674 &    0.070272 &  0.959332 \\
\bottomrule
\end{tabular}
    \label{fig:tab_reg_gb2}
    }
\end{table*}

\begin{table*}[!ht]
    \centering
\scriptsize{
\begin{tabular}{ccccccc}
\toprule
{} &         Dataset Name & Algorithm &   $R^2$ Mean  $\uparrow$  &  MAE   $\downarrow$  &    MSE  $\downarrow$  &   Spearman coeff. $\uparrow$ \\
\midrule
109 &                   feynman\_II\_15\_5 &        GB &  0.998703 &   0.148046 &    0.035469 &  0.999086 \\
110 &                  feynman\_II\_24\_17 &        GB &  0.999786 &   0.009717 &    0.000159 &  0.999882 \\
111 &                  feynman\_II\_27\_16 &        GB &  0.999038 &   2.006489 &     7.70378 &  0.999233 \\
112 &                  feynman\_II\_27\_18 &        GB &   0.99981 &   0.258533 &    0.124766 &  0.999873 \\
113 &                   feynman\_II\_34\_2 &        GB &  0.999065 &   0.224284 &    0.087908 &  0.999368 \\
114 &                 feynman\_II\_34\_29a &        GB &  0.998903 &   0.005328 &    0.000058 &  0.999263 \\
115 &                  feynman\_II\_34\_2a &        GB &   0.99894 &   0.010497 &    0.000223 &  0.999292 \\
\bottomrule
\end{tabular}
    \label{fig:tab_reg_gb3}
    }
\end{table*}

% KNN-------------------------------------------------------------------------------------------------------------------

\begin{table*}[!ht]
    \centering
\scriptsize{
\begin{tabular}{ccccccc}
\toprule
{} &         Dataset Name & Algorithm &   $R^2$ Mean  $\uparrow$  &  MAE   $\downarrow$  &    MSE  $\downarrow$  &   Spearman coeff. $\uparrow$ \\
\midrule
1   &                          1027\_ESL &       KNN &  0.807879 &    0.41613 &     0.37531 &  0.913604 \\
2   &                          1028\_SWD &       KNN &  0.343418 &   0.510123 &    0.424094 &  0.600515 \\
3   &                          1029\_LEV &       KNN &  0.513662 &   0.493256 &    0.439421 &  0.722121 \\
4   &                          1030\_ERA &       KNN &  0.345674 &    1.26361 &    2.507198 &  0.552753 \\
5   &                      1089\_USCrime &       KNN &  0.703708 &  14.772808 &  412.108062 &  0.826278 \\
6   &              1096\_FacultySalaries &       KNN &   0.53663 &   1.953896 &    6.991502 &  0.868185 \\
7   &               1199\_BNG\_echoMonths &       KNN &  0.334282 &  10.376265 &  166.239817 &  0.569137 \\
\bottomrule
\end{tabular}
    \label{fig:tab_reg_knn1}
    }
\end{table*}

\begin{table*}[!ht]
    \centering
\scriptsize{
\begin{tabular}{ccccccc}
\toprule
{} &         Dataset Name & Algorithm &   $R^2$ Mean  $\uparrow$  &  MAE   $\downarrow$  &    MSE  $\downarrow$  &   Spearman coeff. $\uparrow$ \\
\midrule
55  &                 612\_fri\_c1\_1000\_5 &       KNN &  0.889513 &   0.246193 &    0.105308 &  0.942656 \\
56  &                  613\_fri\_c3\_250\_5 &       KNN &  0.826467 &   0.322477 &    0.175191 &  0.893465 \\
57  &                 615\_fri\_c4\_250\_10 &       KNN &  0.530654 &   0.543859 &    0.468519 &  0.708213 \\
58  &                 616\_fri\_c4\_500\_50 &       KNN &  0.210018 &   0.706435 &    0.783715 &  0.476024 \\
59  &                  617\_fri\_c3\_500\_5 &       KNN &  0.859534 &   0.273003 &    0.137508 &  0.922143 \\
60  &                618\_fri\_c3\_1000\_50 &       KNN &  0.311386 &   0.653644 &    0.664081 &  0.557806 \\
61  &                620\_fri\_c1\_1000\_25 &       KNN &  0.375528 &    0.65647 &    0.631844 &  0.608514 \\
\bottomrule
\end{tabular}
    \label{fig:tab_reg_knn2}
    }
\end{table*}

\begin{table*}[!ht]
    \centering
\scriptsize{
\begin{tabular}{ccccccc}
\toprule
{} &         Dataset Name & Algorithm &   $R^2$ Mean  $\uparrow$  &  MAE   $\downarrow$  &    MSE  $\downarrow$  &   Spearman coeff. $\uparrow$ \\
\midrule
109 &                   feynman\_II\_15\_5 &       KNN &  0.999182 &   0.097892 &     0.02238 &  0.999623 \\
110 &                  feynman\_II\_24\_17 &       KNN &  0.999337 &   0.015251 &    0.000493 &   0.99979 \\
111 &                  feynman\_II\_27\_16 &       KNN &  0.999712 &   0.943228 &    2.311306 &  0.999871 \\
112 &                  feynman\_II\_27\_18 &       KNN &   0.99999 &    0.05433 &    0.006645 &  0.999995 \\
113 &                   feynman\_II\_34\_2 &       KNN &  0.999754 &   0.104578 &    0.023089 &  0.999851 \\
114 &                 feynman\_II\_34\_29a &       KNN &  0.999327 &   0.003067 &    0.000035 &  0.999831 \\
115 &                  feynman\_II\_34\_2a &       KNN &  0.999376 &    0.00603 &    0.000131 &  0.999834 \\
\bottomrule
\end{tabular}
    \label{fig:tab_reg_knn3}
    }
\end{table*}

% Linear -------------------------------------------------------------------------------------------------------------------

\begin{table*}[!ht]
    \centering
\scriptsize{
\begin{tabular}{ccccccc}
\toprule
{} &         Dataset Name & Algorithm &   $R^2$ Mean  $\uparrow$  &  MAE   $\downarrow$  &    MSE  $\downarrow$  &   Spearman coeff. $\uparrow$ \\
\midrule
1   &                          1027\_ESL &    Linear &  0.850755 &    0.40513 &     0.292207 &  0.935449 \\
2   &                          1028\_SWD &    Linear &  0.378101 &   0.511349 &     0.401739 &  0.620222 \\
3   &                          1029\_LEV &    Linear &  0.557796 &   0.488484 &     0.399588 &  0.749403 \\
4   &                          1030\_ERA &    Linear &  0.369205 &   1.243903 &     2.418982 &  0.579253 \\
5   &                      1089\_USCrime &    Linear &  0.812045 &  13.428465 &   265.112604 &  0.843669 \\
6   &              1096\_FacultySalaries &    Linear &  0.001744 &   1.531526 &    10.476934 &  0.902399 \\
7   &               1199\_BNG\_echoMonths &    Linear &  0.439689 &   9.255462 &   139.918052 &  0.614086 \\
\bottomrule
\end{tabular}
    \label{fig:tab_reg_linear1}
    }
\end{table*}

\begin{table*}[!ht]
    \centering
\scriptsize{
\begin{tabular}{ccccccc}
\toprule
{} &         Dataset Name & Algorithm &   $R^2$ Mean  $\uparrow$  &  MAE   $\downarrow$  &    MSE  $\downarrow$  &   Spearman coeff. $\uparrow$ \\
\midrule
55  &                 612\_fri\_c1\_1000\_5 &    Linear &  0.229149 &   0.728327 &     0.735948 &  0.470516 \\
56  &                  613\_fri\_c3\_250\_5 &    Linear &  0.263946 &   0.724662 &     0.746828 &  0.505514 \\
57  &                 615\_fri\_c4\_250\_10 &    Linear &  0.276361 &   0.708448 &      0.71733 &  0.471958 \\
58  &                 616\_fri\_c4\_500\_50 &    Linear &  0.205509 &   0.728739 &     0.788715 &  0.453025 \\
59  &                  617\_fri\_c3\_500\_5 &    Linear &  0.216635 &   0.735879 &      0.76765 &  0.449461 \\
60  &                618\_fri\_c3\_1000\_50 &    Linear &  0.265698 &    0.69623 &       0.7071 &  0.472572 \\
61  &                620\_fri\_c1\_1000\_25 &    Linear &  0.250131 &   0.738213 &     0.757734 &  0.478474 \\
\bottomrule
\end{tabular}
    \label{fig:tab_reg_linear2}
    }
\end{table*}

\begin{table*}[!ht]
    \centering
\scriptsize{
\begin{tabular}{ccccccc}
\toprule
{} &         Dataset Name & Algorithm &   $R^2$ Mean  $\uparrow$  &  MAE   $\downarrow$  &    MSE  $\downarrow$  &   Spearman coeff. $\uparrow$ \\
\midrule
109 &                   feynman\_II\_15\_5 &    Linear &  0.210697 &   3.561805 &     21.58674 &  0.430193 \\
110 &                  feynman\_II\_24\_17 &    Linear &  0.970472 &   0.118832 &     0.021962 &  0.997035 \\
111 &                  feynman\_II\_27\_16 &    Linear &  0.802556 &  29.222483 &  1581.939504 &  0.983116 \\
112 &                  feynman\_II\_27\_18 &    Linear &  0.878876 &   6.927826 &    79.418532 &  0.988165 \\
113 &                   feynman\_II\_34\_2 &    Linear &  0.866166 &   2.617153 &    12.578403 &  0.981266 \\
114 &                 feynman\_II\_34\_29a &    Linear &  0.786954 &   0.072809 &     0.011172 &  0.979481 \\
115 &                  feynman\_II\_34\_2a &    Linear &  0.784677 &   0.146543 &     0.045263 &  0.979331 \\
\bottomrule
\end{tabular}
    \label{fig:tab_reg_linear3}
    }
\end{table*}

% Neural Net -------------------------------------------------------------------------------------------------------------------

\begin{table*}[!ht]
    \centering
\scriptsize{
\begin{tabular}{ccccccc}
\toprule
{} &         Dataset Name & Algorithm &   $R^2$ Mean  $\uparrow$  &  MAE   $\downarrow$  &    MSE  $\downarrow$  &   Spearman coeff. $\uparrow$ \\
\midrule
1   &                          1027\_ESL &  NeuralNet &  0.852823 &   0.410395 &  2.893702e-01 &  0.937654 \\
2   &                          1028\_SWD &  NeuralNet &  0.366961 &   0.506931 &  4.093295e-01 &  0.617072 \\
3   &                          1029\_LEV &  NeuralNet &  0.545263 &   0.484831 &  4.109138e-01 &  0.742203 \\
4   &                          1030\_ERA &  NeuralNet &  0.365520 &   1.244535 &  2.433973e+00 &  0.568572 \\
5   &                      1089\_USCrime &  NeuralNet &  0.740107 &  15.647898 &  3.739526e+02 &  0.804915 \\
6   &              1096\_FacultySalaries &  NeuralNet &  0.784680 &   1.450788 &  3.476780e+00 &  0.883562 \\
7   &               1199\_BNG\_echoMonths &  NeuralNet &  0.442570 &   9.068394 &  1.392056e+02 &  0.631720 \\
\bottomrule
\end{tabular}
    \label{fig:tab_reg_nn1}
    }
\end{table*}

\begin{table*}[!ht]
    \centering
\scriptsize{
\begin{tabular}{ccccccc}
\toprule
{} &         Dataset Name & Algorithm &   $R^2$ Mean  $\uparrow$  &  MAE   $\downarrow$  &    MSE  $\downarrow$  &   Spearman coeff. $\uparrow$ \\
\midrule
55  &                 612\_fri\_c1\_1000\_5 &  NeuralNet &  0.901813 &   0.232382 &  9.276487e-02 &  0.951857 \\
56  &                  613\_fri\_c3\_250\_5 &  NeuralNet &  0.687361 &   0.412556 &  3.149878e-01 &  0.831824 \\
57  &                 615\_fri\_c4\_250\_10 &  NeuralNet &  0.497194 &   0.537168 &  4.991084e-01 &  0.693061 \\
58  &                 616\_fri\_c4\_500\_50 &  NeuralNet &  0.091603 &   0.764015 &  9.019947e-01 &  0.375784 \\
59  &                  617\_fri\_c3\_500\_5 &  NeuralNet &  0.769612 &   0.338511 &  2.308483e-01 &  0.885101 \\
60  &                618\_fri\_c3\_1000\_50 &  NeuralNet &  0.198645 &   0.696363 &  7.710063e-01 &  0.445806 \\
61  &                620\_fri\_c1\_1000\_25 &  NeuralNet &  0.242152 &   0.686101 &  7.655733e-01 &  0.523801 \\
\bottomrule
\end{tabular}
    \label{fig:tab_reg_nn2}
    }
\end{table*}

\begin{table*}[!ht]
    \centering
\scriptsize{
\begin{tabular}{ccccccc}
\toprule
{} &         Dataset Name & Algorithm &   $R^2$ Mean  $\uparrow$  &  MAE   $\downarrow$  &    MSE  $\downarrow$  &   Spearman coeff. $\uparrow$ \\
\midrule
109 &                   feynman\_II\_15\_5 &  NeuralNet &  0.999063 &   0.117654 &  2.562177e-02 &  0.999707 \\
110 &                  feynman\_II\_24\_17 &  NeuralNet &  0.999730 &   0.010473 &  2.008074e-04 &  0.999984 \\
111 &                  feynman\_II\_27\_16 &  NeuralNet &  0.999488 &   1.392176 &  4.111538e+00 &  0.999896 \\
112 &                  feynman\_II\_27\_18 &  NeuralNet &  0.999613 &   0.360378 &  2.543454e-01 &  0.999967 \\
113 &                   feynman\_II\_34\_2 &  NeuralNet &  0.999546 &   0.147771 &  4.276658e-02 &  0.999939 \\
114 &                 feynman\_II\_34\_29a &  NeuralNet &  0.999319 &   0.003836 &  3.560954e-05 &  0.999918 \\
115 &                  feynman\_II\_34\_2a &  NeuralNet &  0.999368 &   0.008007 &  1.326014e-04 &  0.999930 \\
\bottomrule
\end{tabular}
    \label{fig:tab_reg_nn3}
    }
\end{table*}

% Random Forest -------------------------------------------------------------------------------------------------------------------

\begin{table*}[!ht]
    \centering
\scriptsize{
\begin{tabular}{ccccccc}
\toprule
{} &         Dataset Name & Algorithm &   $R^2$ Mean  $\uparrow$  &  MAE   $\downarrow$  &    MSE  $\downarrow$  &   Spearman coeff. $\uparrow$ \\
\midrule
1   &                          1027\_ESL &        RF &  0.828191 &   0.401064 &    0.337169 &  0.926962 \\
2   &                          1028\_SWD &        RF &  0.351163 &   0.509689 &    0.419149 &  0.603557 \\
3   &                          1029\_LEV &        RF &  0.513733 &   0.494058 &    0.439321 &  0.721755 \\
4   &                          1030\_ERA &        RF &  0.344498 &   1.263994 &    2.511468 &  0.551032 \\
5   &                      1089\_USCrime &        RF &  0.787284 &  14.124231 &  307.610681 &  0.825673 \\
6   &              1096\_FacultySalaries &        RF &  0.646406 &   1.674051 &    5.056379 &   0.86207 \\
7   &               1199\_BNG\_echoMonths &        RF &  0.466759 &   8.845958 &  133.148714 &  0.658935 \\
\bottomrule
\end{tabular}
    \label{fig:tab_reg_rf1}
    }
\end{table*}

\begin{table*}[!ht]
    \centering
\scriptsize{
\begin{tabular}{ccccccc}
\toprule
{} &         Dataset Name & Algorithm &   $R^2$ Mean  $\uparrow$  &  MAE   $\downarrow$  &    MSE  $\downarrow$  &   Spearman coeff. $\uparrow$ \\
\midrule
55  &                 612\_fri\_c1\_1000\_5 &        RF &  0.922967 &   0.208122 &    0.073467 &  0.958037 \\
56  &                  613\_fri\_c3\_250\_5 &        RF &  0.849498 &   0.301571 &    0.152947 &  0.906948 \\
57  &                 615\_fri\_c4\_250\_10 &        RF &   0.76537 &    0.36884 &    0.235968 &  0.871497 \\
58  &                 616\_fri\_c4\_500\_50 &        RF &  0.835236 &   0.309777 &     0.16219 &   0.90138 \\
59  &                  617\_fri\_c3\_500\_5 &        RF &  0.879998 &   0.255318 &    0.117882 &  0.933057 \\
60  &                618\_fri\_c3\_1000\_50 &        RF &  0.883478 &   0.264537 &    0.112224 &  0.923763 \\
61  &                620\_fri\_c1\_1000\_25 &        RF &  0.874473 &   0.282563 &    0.126939 &  0.931109 \\
\bottomrule
\end{tabular}
    \label{fig:tab_reg_rf2}
    }
\end{table*}

\begin{table*}[!ht]
    \centering
\scriptsize{
\begin{tabular}{ccccccc}
\toprule
{} &         Dataset Name & Algorithm &   $R^2$ Mean  $\uparrow$  &  MAE   $\downarrow$  &    MSE  $\downarrow$  &   Spearman coeff. $\uparrow$ \\
\midrule
109 &                   feynman\_II\_15\_5 &        RF &  0.999697 &   0.062022 &     0.00828 &  0.999894 \\
110 &                  feynman\_II\_24\_17 &        RF &  0.999885 &   0.006752 &    0.000086 &  0.999944 \\
111 &                  feynman\_II\_27\_16 &        RF &   0.99971 &   0.891547 &    2.322599 &  0.999906 \\
112 &                  feynman\_II\_27\_18 &        RF &  0.999992 &   0.044954 &    0.005057 &  0.999997 \\
113 &                   feynman\_II\_34\_2 &        RF &  0.999737 &   0.104984 &    0.024754 &  0.999887 \\
114 &                 feynman\_II\_34\_29a &        RF &  0.999531 &   0.002667 &    0.000025 &  0.999868 \\
115 &                  feynman\_II\_34\_2a &        RF &  0.999604 &   0.005112 &    0.000083 &  0.999869 \\
\bottomrule
\end{tabular}
    \label{fig:tab_reg_rf3}
    }
\end{table*}

% LGBM -------------------------------------------------------------------------------------------------------------------

\begin{table*}[!ht]
    \centering
\scriptsize{
\begin{tabular}{ccccccc}
\toprule
{} &         Dataset Name & Algorithm &   $R^2$ Mean  $\uparrow$  &  MAE   $\downarrow$  &    MSE  $\downarrow$  &   Spearman coeff. $\uparrow$ \\
\midrule
1   &                          1027\_ESL &      LGBM &  0.844445 &   0.413833 &    0.305865 &  0.931222 \\
2   &                          1028\_SWD &      LGBM &  0.359984 &   0.506554 &    0.413474 &  0.609287 \\
3   &                          1029\_LEV &      LGBM &  0.524891 &   0.490583 &    0.429172 &  0.728595 \\
4   &                          1030\_ERA &      LGBM &  0.346505 &   1.262428 &    2.504109 &  0.552466 \\
5   &                      1089\_USCrime &      LGBM & -0.160501 &   36.23973 &  1727.36776 &       NaN \\
6   &              1096\_FacultySalaries &      LGBM & -0.103898 &   3.356827 &   20.915339 &  0.760385 \\
7   &               1199\_BNG\_echoMonths &      LGBM &  0.476546 &   8.791663 &  130.713328 &  0.666369 \\
\bottomrule
\end{tabular}
    \label{fig:tab_reg_lgbm1}
    }
\end{table*}

\begin{table*}[!ht]
    \centering
\scriptsize{
\begin{tabular}{ccccccc}
\toprule
{} &         Dataset Name & Algorithm &   $R^2$ Mean  $\uparrow$  &  MAE   $\downarrow$  &    MSE  $\downarrow$  &   Spearman coeff. $\uparrow$ \\
\midrule
55  &                 612\_fri\_c1\_1000\_5 &      LGBM &  0.943877 &   0.178135 &    0.053381 &  0.969261 \\
56  &                  613\_fri\_c3\_250\_5 &      LGBM &  0.854812 &   0.281067 &    0.147232 &   0.91031 \\
57  &                 615\_fri\_c4\_250\_10 &      LGBM &  0.815332 &   0.309522 &    0.188706 &  0.895958 \\
58  &                 616\_fri\_c4\_500\_50 &      LGBM &  0.856199 &   0.277982 &    0.140655 &  0.908752 \\
59  &                  617\_fri\_c3\_500\_5 &      LGBM &  0.895849 &   0.225265 &    0.102203 &  0.944819 \\
60  &                618\_fri\_c3\_1000\_50 &      LGBM &  0.915046 &   0.217885 &    0.081729 &  0.942882 \\
61  &                620\_fri\_c1\_1000\_25 &      LGBM &  0.928287 &   0.209569 &    0.072189 &  0.959867 \\
\bottomrule
\end{tabular}
    \label{fig:tab_reg_lgbm2}
    }
\end{table*}

\begin{table*}[!ht]
    \centering
\scriptsize{
\begin{tabular}{ccccccc}
\toprule
{} &         Dataset Name & Algorithm &   $R^2$ Mean  $\uparrow$  &  MAE   $\downarrow$  &    MSE  $\downarrow$  &   Spearman coeff. $\uparrow$ \\
\midrule
109 &                   feynman\_II\_15\_5 &      LGBM &  0.998989 &   0.125219 &    0.027641 &  0.999477 \\
110 &                  feynman\_II\_24\_17 &      LGBM &  0.999691 &   0.011978 &     0.00023 &  0.999832 \\
111 &                  feynman\_II\_27\_16 &      LGBM &  0.999096 &   1.917405 &    7.245248 &  0.999355 \\
112 &                  feynman\_II\_27\_18 &      LGBM &  0.999833 &   0.246181 &    0.109237 &  0.999893 \\
113 &                   feynman\_II\_34\_2 &      LGBM &  0.998982 &   0.234243 &    0.095705 &  0.999271 \\
114 &                 feynman\_II\_34\_29a &      LGBM &  0.998847 &   0.005464 &     0.00006 &  0.999204 \\
115 &                  feynman\_II\_34\_2a &      LGBM &  0.998851 &   0.010951 &    0.000242 &  0.999194 \\
\bottomrule
\end{tabular}
    \label{fig:tab_reg_lgbm3}
    }
\end{table*}

\begin{table*}[!ht]
\section{Detailed Results for Anomaly Detection}
% \vskip 0.1in
Our method Familiarity Conviction (FC) and Similarity conviction (SC) is compared with six other popular methods for carrying out anomaly detection. DeepSVDD was trained on 20 epochs with the inlier set of training data. Picking a conviction level of 0.7 for all datasets (wihout choosing it in a dataset specific manner), our method achieves the highest $F1$ scores in 12 of the 20 datasets.

    \centering
    \scriptsize{
    \caption{Information of the 20 ODDS dataset used for anomaly detection}
    \vskip 0.1in
    \begin{tabular}{|c|c|c|}
    \hline
        Dataset & Dataset Size & \% Anomalies \\ \hline
        wine.mat & [129, 13] & 7.70\% \\ \hline
        wbc.mat & [278, 30] & 5.60\% \\ \hline
        vowels.mat & [456, 12] & 3.40\% \\ \hline
        vertebral.mat & [240, 6] & 12.50\% \\ \hline
        thyroid.mat & [3772, 6] & 2.50\% \\ \hline
        speech.mat & [3686, 400] & 1.65\% \\ \hline
        shuttle.mat & [49097, 9] & 7\% \\ \hline
        satimage-2.mat & [5803, 36] & 1.20\% \\ \hline
        satellite.mat & [6435, 36] & 32\% \\ \hline
        pima.mat & [768, 8] & 35\% \\ \hline
        optdigits.mat & [5216, 64] & 3\% \\ \hline
        musk.mat & [3062, 166] & 3.20\% \\ \hline
        mnist.mat & [7603, 100] & 9.20\% \\ \hline
        lympho.mat & [148, 18] & 4.10\% \\ \hline
        letter.mat & [1600, 32] & 6.25\% \\ \hline
        ionosphere.mat & [351, 33] & 36\% \\ \hline
        glass.mat & [214, 9] & 4.20\% \\ \hline
        cardio.mat & [1831, 21] & 9.60\% \\ \hline
        breastw.mat & [683, 9] & 35\% \\ \hline
        arrhythmia.mat &	[452, 274] &	15 \% \\ \hline
    \end{tabular}
    \label{fig:anomaly_detection_data_info}
    }
\end{table*}

\begin{table*}[!ht]
    \centering
     \caption{F1 scores of Anomaly Detection on ODDS datasets.} \text{(\color{blue}{\textbf{Blue}} values indicate the best performance; \color{brown}{\textbf{Brown}} values indicate the second-best performance )}
     % Best scores for each dataset are in \textbf{bold}.}
     \vskip 0.1in
    \begin{tabular}{|c|c|c|c|c|c|c|c|c|}
    \hline
    Dataset & \textbf{Ours (FC)} & \textbf{Ours (SC)} & OCSVM  & IForest & CBLOF         & LOF & ECOD  &  DeepSVDD \\ \hline
    wine         & 0.44                          & 0.18                         & 0.31 & 0.1             & \color{brown}\textbf{0.87}          & \color{blue}\textbf{0.95}      & 0.24 & 0.53\\ \hline
    wbc          & \color{brown}\textbf{0.57}                          & \color{blue}\textbf{0.65}                & 0.54 & 0.61            & 0.51          & 0.40               & 0.44  & 0.49 \\ \hline
    vowels       & 0.20                          & \color{blue}\textbf{0.75}                & 0.21 & 0.21            & 0.37          & 0.10               & 0.17 & \color{brown}\textbf{0.42} \\ \hline
    vertebral    & \color{blue}\textbf{0.28}                 & \color{brown}\textbf{0.19}                         & 0.05 & 0.04            & 0.04          & 0.00               & 0.14 & 0.05 \\ \hline
    thyroid      & 0.26                          & \color{blue}\textbf{0.64}                & 0.29 & 0.54            & 0.30          & 0.19               & \color{brown}\textbf{0.56}  & 0.33 \\ \hline
    speech       & \color{brown}\textbf{0.06}                          & \color{blue}\textbf{0.10}                & 0.02 & 0.00            & 0.03          & 0.00               & \color{brown}\textbf{0.06} & \color{brown}\textbf{0.06} \\ \hline
    shuttle      & 0.26                          & 0.60                         & 0.33 & \color{blue}\textbf{0.89}   & \color{brown}\textbf{0.82}          & 0.10               & 0.75 & 0.60 \\ \hline
    satimage-2   & 0.10                          & \color{blue}\textbf{0.94}                & 0.39 & \color{brown}\textbf{0.41}            & 0.22          & 0.14               & 0.34 & 0.26 \\ \hline
    satellite    & 0.58                          & \color{blue}\textbf{0.75}                & 0.14 & 0.51            & 0.48          & 0.04               & 0.29 & \color{brown}\textbf{0.66} \\ \hline
    pima         & \color{blue}\textbf{0.52}                 & 0.01                         & 0.12 & \color{brown}\textbf{0.29}            & 0.25          & 0.06               & 0.22 & \color{blue}\textbf{0.52} \\ \hline
    optdigits    & 0.10                          & 0.00                         & 0.03 & 0.08            & \color{brown}\textbf{0.18} & 0.00               & 0.03 & \color{blue}\textbf{0.26} \\ \hline
    musk         & 0.22                          & \color{blue}\textbf{0.78}                & 0.14 & \color{brown}\textbf{0.71}            & 0.48          & 0.00               & 0.20 & 0.48 \\ \hline
    mnist        & 0.34                          & 0.24                         & 0.19 & \color{brown}\textbf{0.39}   & 0.34          & 0.01               & 0.20 & \color{blue}\textbf{0.43} \\ \hline
    lympho       & 0.35                          & \color{blue}\textbf{0.73}                & \color{brown}\textbf{0.38} & 0.36            & 0.29          & 0.00               & 0.09 & 0.26 \\ \hline
    letter       & 0.17                          & \color{blue}\textbf{0.43}                & 0.18 & 0.10            & 0.22          & 0.08               & 0.13 & \color{brown}\textbf{0.31}\\ \hline
    ionosphere   & 0.59                          & \color{brown}\textbf{0.85}                & 0.27 & 0.67            & 0.43          & 0.78               & 0.32 & \color{blue}\textbf{0.90} \\ \hline
    glass        & \color{brown}\textbf{0.20}                          & 0.14                         & 0.13 & 0.14            & 0.13          & \color{blue}\textbf{0.30}      & 0.19 & \color{blue}\textbf{0.30} \\ \hline
    cardio       & 0.35                          & 0.51                         & 0.24 & 0.44            & \color{brown}\textbf{0.54}  & 0.10               & 0.48 & \color{blue}\textbf{0.61} \\ \hline
    breastw      & 0.33                          & 0.86                         & 0.17 & \color{brown}\textbf{0.90}   & 0.42          & 0.09               & 0.32 & \color{blue}\textbf{0.94} \\ \hline
    arrhythmia   & \color{brown}\textbf{0.47}                          & \color{blue}\textbf{0.51}                & 0.23 & 0.11            & 0.41          & 0.37               & 0.43 & \color{blue}\textbf{0.51} \\ \hline
    \end{tabular}
    \label{fig:anomaly_detection_results_full}
\end{table*}

%%%%%%%%%%%%%%%%%%%%%%%%%%%%%%%%%%%%%%%%%%%%%%%%%%%%%%%%%%%%%%%%%%%%%%%%%%%%%%%
%%%%%%%%%%%%%%%%%%%%%%%%%%%%%%%%%%%%%%%%%%%%%%%%%%%%%%%%%%%%%%%%%%%%%%%%%%%%%%%

\end{document}